\documentclass[letterpaper,table,dvipsnames]{article}
\usepackage{aaai2026}  
\usepackage{times}  
\usepackage{helvet}  
\usepackage{courier}  
\usepackage[hyphens]{url}  
\usepackage{graphicx} 
\usepackage{booktabs}
\usepackage{multirow}
\usepackage{natbib}  
\usepackage{caption} 
\usepackage{algorithm}
\usepackage{algorithmic}
\usepackage{cuted}  

\usepackage{amssymb} 
\usepackage{booktabs} 
\usepackage{amsmath}
\usepackage{graphicx} 
\usepackage{pifont}  
\usepackage{array}    
\definecolor{grayrow}{gray}{0.92}
\usepackage{tabularx}  
\usepackage[most]{tcolorbox}
\usepackage{graphicx}
\usepackage{lipsum}  
\usepackage{enumitem}

\usepackage{newfloat}
\usepackage{listings}
\DeclareCaptionStyle{ruled}{labelfont=normalfont,labelsep=colon,strut=off} 
\floatstyle{ruled}
\newfloat{listing}{tb}{lst}{}
\floatname{listing}{Listing}
\title{\textit{DeepResearch Arena}: The First Exam of LLMs’ Research Abilities via Seminar-Grounded Tasks}

\author{
    Haiyuan Wan \textsuperscript{\rm 1,\rm 2}  \hspace{-0.2em} \thanks{These authors contributed equally to this work.},
    Chen Yang \textsuperscript{\rm 3} \hspace{-0.2em} \footnotemark[1],
    Junchi Yu \textsuperscript{\rm 4},
    Meiqi Tu \textsuperscript{\rm 5},
    Jiaxuan Lu \textsuperscript{\rm 1},
    Di Yu \textsuperscript{\rm 1,\hspace{-0.2em} \rm 2},
    Jianbao Cao \textsuperscript{\rm 1,\hspace{-0.2em} \rm 6},
    Ben Gao \textsuperscript{\rm 1, \rm 6},
    Jiaqing Xie \textsuperscript{\rm 1},
    Aoran Wang \textsuperscript{\rm 1},
    Wenlong Zhang \textsuperscript{\rm 1},
    Philip Torr \textsuperscript{\rm 4},
    Dongzhan Zhou \textsuperscript{\rm 1}\thanks{Corresponding author}
}

\affiliations{
    \textsuperscript{\rm 1}Shanghai Artificial Intelligence Laboratory \\
    \textsuperscript{\rm 2}Tsinghua University \quad
    \textsuperscript{\rm 3}The Hong Kong University of Science and Technology, Guangzhou \quad
    \textsuperscript{\rm 4}University of Oxford \quad \\
    \textsuperscript{\rm 5}The University of Hong Kong \quad 
    \textsuperscript{\rm 6}Wuhan University \quad

}

\usepackage{bibentry}

\begin{document}
\maketitle

\begin{figure*}[t]
  \centering
  \includegraphics[width=\linewidth]{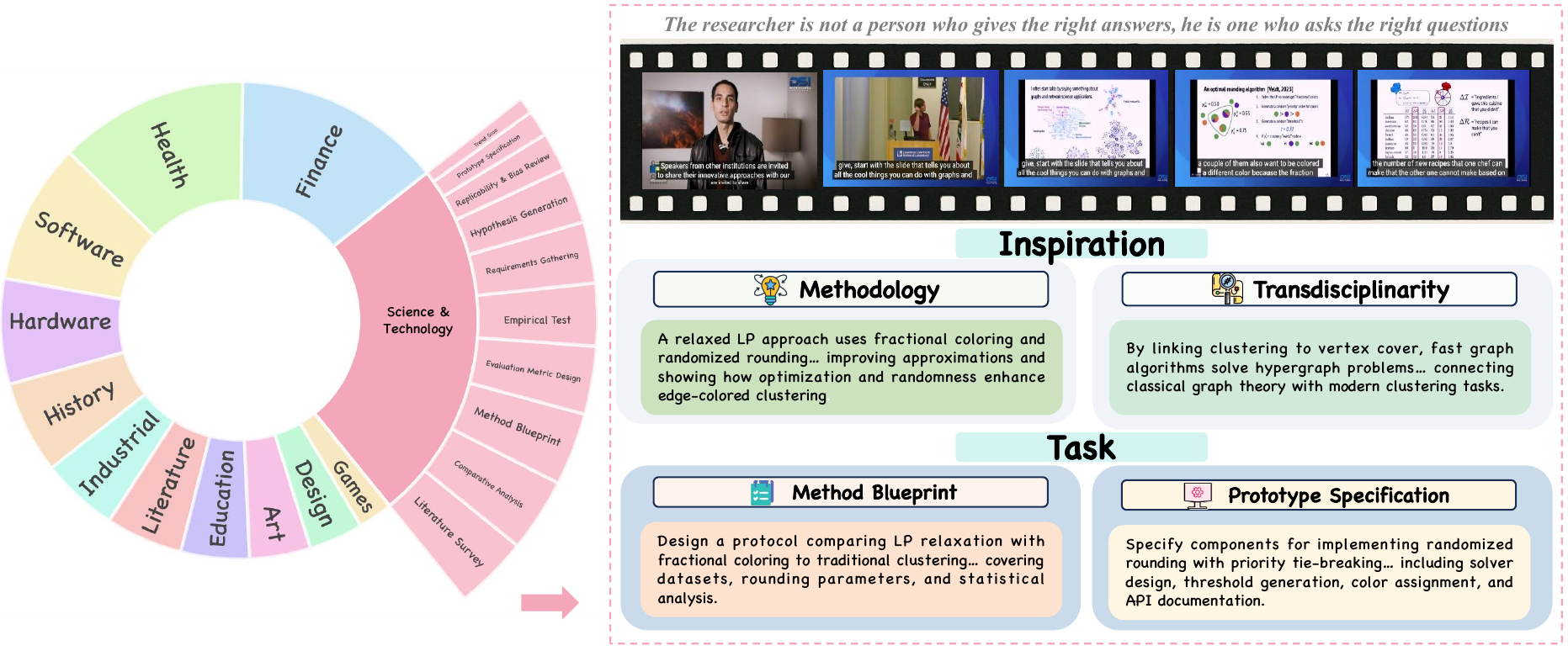}
  \caption{\textbf{Overview of seminar domains and task structures in MAHTG.} 
\textbf{Left:} Distribution of academic seminars across diverse domains such as Science \& Technology, Health, Finance, and others. The outer arc further decomposes each domain into representative research tasks. For instance, Science \& Technology includes tasks such as \textit{Hypothesis Generation}, \textit{Empirical Test}, \textit{Prototype Specification}, and \textit{Trend Scan}.
\textbf{Right:} Illustration of MAHTG's multi-agent pipeline, where seminar content is transformed into structured research tasks via intermediate inspirations (e.g., \textit{Methodology}, \textit{Transdisciplinarity}). Example outputs are shown for both stages.}
  \label{fig:overview}
\end{figure*}

\begin{abstract}
Deep research agents have attracted growing attention for their potential to orchestrate multi-stage research workflows, spanning literature synthesis, methodological design, and empirical verification. 
Despite these strides, evaluating their research capability faithfully is rather challenging due to the difficulty of collecting frontier research questions that genuinely capture researchers’ attention and intellectual curiosity.
To address this gap, we introduce \textit{\textbf{DeepResearch Arena}}, a benchmark grounded in academic seminars that capture rich expert discourse and interaction, better reflecting real-world research environments and reducing the risk of data leakage.  
To automatically construct DeepResearch Arena, we propose a Multi-Agent Hierarchical Task Generation (MAHTG) system that extracts research-worthy inspirations from seminar transcripts. 
The MAHTG system further translates research-worthy inspirations into high-quality research tasks, ensuring the traceability of research task formulation while filtering noise. 
With the MAHTG system, we curate DeepResearch Arena with over 10,000 high-quality research tasks from over 200 academic seminars, spanning 12 disciplines, such as literature, history, and science. 
Our extensive evaluation shows that DeepResearch Arena presents substantial challenges for current state-of-the-art agents, with clear performance gaps observed across different models.

\end{abstract}



\section{Introduction}
Recent developments in large language models (LLMs) have led to the rise of the deep research agent ~\cite{huang2025deepresearchagentssystematic,xu2025comprehensivesurveydeepresearch, wu2025agenticreasoningstreamlinedframework}, a LLM-powered agentic system designed for research task automation by integrating literature search~\cite{baek2024researchagent}, experiment design~\cite{schmidgall2025agentlaboratoryusingllm}, and ideation~\cite{li2024chain}.
Prevailing examples, such as GPT DeepResearch~\cite{openai2025deepresearch}, indicate that deep research agents have great potential to significantly promote research creativity and productivity.


While deep research agents have gained increasing attention~\cite{du2025deepresearch}, faithfully evaluating their research ability remains a huge challenge.
As Einstein once stated, \textit{The formulation of the problem is often more essential than its solution, which may be merely a matter of mathematical or experimental skill}~\cite{einstein1938evolution}.
This perspective highlights a crucial challenge in formulating high-quality and frontier research tasks to faithfully assess the ability of deep research agents.

Existing benchmarks for deep research agents mainly resort to two approaches to acquire research questions. The first leverages static data corpora such as academic literature and web content, as seen in AcademicBrowse~\cite{zhou2025academicbrowse}, BrowseComp~\cite{wei2025browsecomp}, and Researchbench~\cite{liu2025researchbench}. The second approach involves manually curated research tasks by domain experts, exemplified by Humanity’s Last Exam~\cite{phan2025humanitysexam}, DeepResearchBench~\cite{du2025deepresearch}, and ExpertLongBench~\cite{ruan2025expertlongbench}. However, both approaches are hindered by critical limitations. Benchmarks derived from static corpora risk data leakage, as the underlying content may already be included in the model pertaining. Meanwhile, datasets curated by experts face scalability bottlenecks and often lack the diversity and spontaneity found in authentic research settings. More fundamentally, both sources tend to abstract away from the situated, evolving nature of real-world research inquiry, where questions emerge dynamically through discourse, ambiguity, and interdisciplinary exploration. A detailed comparison of these benchmarks across key dimensions, including scalability, automation, data leakage risk, and research realism, is provided in Table~\ref{tab:benchmark_comparison}.

To bridge this gap, we introduce a novel benchmark, \textit{\textbf{Deep Research Arena}}, designed to evaluate deep research agents under authentic, cognitively demanding research scenarios. Unlike static corpora that present information without context, or expert-curated benchmarks that rely on handcrafted tasks detached from actual discovery processes, the proposed benchmark is grounded in academic seminars, where real researchers pose open-ended questions, explore uncertain ideas, and build shared understanding through live discussion. This source captures how real research problems naturally emerge, making Deep Research Arena a more faithful proxy of real-world inquiry. Furthermore, seminar videos are rarely included in model pretraining, which significantly reduces the risk of data leakage that commonly affects benchmarks derived from literature or web corpora.


To capture the nature of such authentic inquiry, Deep Research Arena formulates tasks as open-ended, under-defined problems, drawn by the theory of Ill-Structured Problem Solving~\cite{jonassen1997isp}, which describes real-world problems as “poorly defined, with no single correct formulation and no objective evaluation criteria”. To construct the Deep Research Arena, we develop a Multi-Agent Hierarchical Task Generation (MAHTG) system that automatically extracts research-worthy inspirations and systematically transforms them into high-quality, traceable research tasks through a multi-stage filtering and structuring pipeline. This design ensures both the authenticity and reproducibility of task construction, while reducing noise and preserving the intellectual context of original expert discourse.

We curate a large-scale, multidisciplinary seminar dataset, constructing over 10,000 structured tasks spanning core research competencies. Building on this, we develop a hybrid evaluation framework that jointly measures factual grounding and higher-order reasoning, with examples shown in Figure~\ref{fig:overview}. Together, these contributions provide a rigorous and theory-aligned foundation for assessing deep research competence in deep research agents.

Our contributions are threefold:
\begin{itemize}
\item \textbf{Seminar-grounded data collection.} We curate a corpus of over 200 academic seminars across 12 disciplines, encompassing real-world expert discourse across science, engineering, humanities, and the arts.

\item \textbf{Hierarchical task generation.} A multi-stage agent framework extracts research-worthy inspirations from seminar transcripts, categorized into \textit{Limitation}, \textit{Methodology}, \textit{Transdisciplinarity}, and \textit{Hypothesis}, and transforms them into over 10{,}000 open-ended tasks aligned with the canonical research stages of \textit{Synthesis}, \textit{Design}, and \textit{Evaluation}.

\item \textbf{Hybrid evaluation framework.} We employ two complementary metrics to quantify factual alignment via extracted keypoints and evaluate open-ended reasoning using adaptively generated, rubric-based checklists.
\end{itemize}

\begin{table*}[!t]
\centering
\small
\renewcommand{\arraystretch}{1.15}
\setlength{\tabcolsep}{6pt}
\begin{tabular}{lccccc}
\toprule
\textbf{Benchmark} & \textbf{Data Source} & \textbf{Scalability} & \textbf{Risk of Data Leakage} & \textbf{Task Automation} & \textbf{Research Realism} \\
\midrule
ScholarSearch       & Literature        & \ding{51} & \ding{51} & \ding{55} & \ding{55} \\
BrowseComp              & Web Corpus        & \ding{51} & \ding{51} & \ding{51} & \ding{55} \\
ResearchBench         & Literature        & \ding{51} & \ding{51} & \ding{51} & \ding{55} \\
\midrule
Humanity’s Last Exam & Expert   & \ding{55} & \ding{51} & \ding{55} & \ding{51} \\
DeepResearchBench       & Expert   & \ding{55} & \ding{51} & \ding{55} & \ding{51} \\
ExpertLongBench    & Expert   & \ding{55} & \ding{51} & \ding{55} & \ding{51} \\
\midrule
\textit{\textbf{DeepResearch Arena (Ours)}}                & Seminar Discourse & \ding{51} & \ding{55} & \ding{51} & \ding{51} \\
\bottomrule
\end{tabular}
\caption{Comparison of existing deep research benchmarks and our \textit{\textbf{DeepResearch Arena}} along key dimensions.}
\label{tab:benchmark_comparison}
\end{table*}

\section{Related Works}
\paragraph{Deep Research Agents.}
The emergence of DR agents builds upon recent advances in LLMs equipped with tool-use capabilities~\cite{li2025adaptivetooluselarge, Qu_2025, tang2023toolalpacageneralizedtoollearning}, which allow models to interface with search engines, code interpreters, and external APIs to extend their reasoning horizon. On this foundation, systems such as GPT Deep Research~\cite{openai2025deepresearch}, Gemini Deep Research~\cite{google2025deepresearch}, and Grok DeepSearch~\cite{xai2025grok3} have been developed to support multi-stage research workflows.
GPT’s system focuses on outline-driven long-form synthesis with citation grounding, Gemini emphasizes multimodal retrieval and synthesis, while Grok prioritizes real-time web summarization for dynamic topics. These agents reflect a shift from retrieval-based assistants to goal-directed, tool-augmented agents capable of supporting exploratory, open-ended inquiry \cite{yu2023thought}.

\paragraph{Benchmarks for Deep Research Agents.}
Existing benchmarks for deep research agents mainly resort to two approaches to acquire research questions: automatically deriving tasks from static corpora or manually curating them through expert design. The first leverages static data corpora such as papers, and web documents to construct benchmarks represented by multi-hop reasoning or simplified scientific queries. Examples include MuSiQue~\cite{trivedi2022musiquemultihopquestionssinglehop}, which automatically generates multi-hop questions by linking single-hop QA pairs from existing datasets, and HotpotQA~\cite{yang2018hotpotqa}, where annotators write questions guided by system-selected Wikipedia article pairs, making the process closer to extraction than genuine question generation. Other benchmarks in this category include StrategyQA~\cite{geva-etal-2021-aristotle}, ThoughtSource~\cite{Ott_2023}, AcademicBrowse~\cite{zhou2025academicbrowse}, and BrowseComp~\cite{wei2025browsecomp}.
Despite their emphasis on multi-step reasoning, these benchmarks rely on manually constructed logic chains with predefined paths. They primarily test factual retrieval and compositional reasoning capabilities, yet fail to capture how research questions naturally emerge, evolve, and iterate in real-world research contexts.
ScienceQA~\cite{lu2022learnexplainmultimodalreasoning} is a large-scale multimodal multiple-choice science QA benchmark (~21K questions across STEM and social/language science) that includes lecture and explanation‑level CoT annotations to support interpretable multi-step reasoning.

\begin{figure*}[!t]
  \centering
  \includegraphics[width=\linewidth]{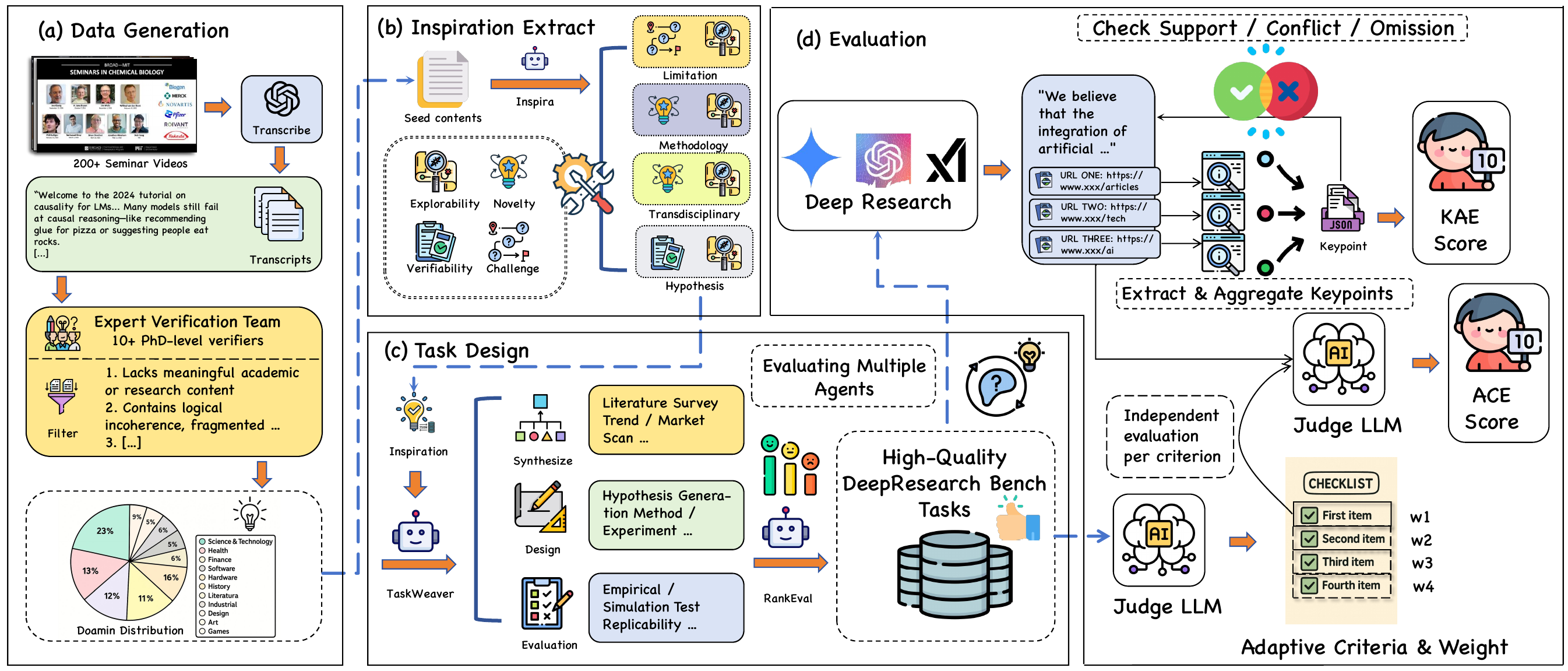}
  \caption{Overview of our benchmark construction pipeline, including four stages: (a) Data generation from transcribed seminar videos, (b) extraction of research inspirations, (c) multi-phase task design, and (d) evaluation using both KAE and ACE metrics.}
  \label{fig:main}
\end{figure*}

The second category consists of expert-authored benchmarks, where researchers collaborate with domain specialists to construct high-quality, PhD-level evaluation tasks. Compared to benchmarks built from static corpora, these datasets typically feature more original, conceptually challenging, and discipline-specific questions that better reflect expert-level reasoning. Representative examples include LAB-Bench~\cite{laurent2024labbenchmeasuringcapabilitieslanguage}, ARC~\cite{clark2018think}, GPQA~\cite{rein2024gpqa}, FrontierMath~\cite{glazer2024frontiermathbenchmarkevaluatingadvanced}, and Humanity’s Last Exam~\cite{phan2025humanitysexam}.
GPQA provides graduate-level multiple-choice questions in biology, physics, and chemistry, curated and verified by domain PhDs to ensure they cannot be solved via surface-level heuristics or web search.
Humanity’s Last Exam comprises a collection of open-ended, expert-written research questions across disciplines such as history, philosophy, and theoretical science, designed to probe creative, integrative thinking under minimal structural constraints. DeepResearch Bench~\cite{du2025deepresearch} moves toward more realistic simulation by requiring long-form research reports across disciplines. However, this entire class of expert-authored benchmarks faces several limitations: their prompts are manually constructed, which restricts scalability and diversity, and the datasets remain relatively small in size. More fundamentally, they also fail to capture how research questions emerge dynamically through discourse, ambiguity, and interdisciplinary exploration—core characteristics of authentic research practice.

\section{Multi-Agent Hierarchical Task Generation.}

\begin{table*}[t]
\centering
{\small
\setlength{\tabcolsep}{4pt}
\renewcommand{\arraystretch}{1.2}
\begin{tabular}{p{0.14\linewidth} p{0.40\linewidth} p{0.41\linewidth}}
\toprule
\textbf{Term} & \textbf{Illustration} & \textbf{Example} \\
\midrule

\rowcolor{grayrow}
\multicolumn{3}{l}{\textbf{I. Core Unit: Inspiration}} \\

Inspiration & A research-worthy idea distilled from academic discourse, exhibiting at least two of: \textit{novelty}, \textit{explorability}, \textit{challenge}, \textit{verifiability}. Serves as the seed for task generation. & “A greedy maximal independent‑set algorithm … achieves a 2‑approximation in O (sum of hyperedge sizes) time … shows classical graph methods can solve edge‑colored hypergraph clustering without auxiliary graphs.” \\

\rowcolor{grayrow}
\multicolumn{3}{l}{\textbf{II. Types of Inspiration}} \\

Limitation & An open problem, deficiency, or bottleneck in existing methods. & “Few models handle transdisciplinary seminar reasoning.” \\
Methodology & A new or adapted approach, pipeline, or tool. & “Introduce retrieval-augmented reranking framework.” \\
Transdisciplinarity & Ideas involving the fusion of theories or tools across disciplines. & “Apply ecological network theory to social dynamics” \\
Hypothesis & A testable proposition that guides design or evaluation. & “Grounded citations improve factual accuracy.” \\

\rowcolor{grayrow}
\multicolumn{3}{l}{\textbf{III. Task Phase Labels}} \\

Synthesize & Collecting, integrating, and analyzing prior work to form direction. & “Identify gaps in seminar-based QA literature.” \\
Design & Designing solutions, models, or experiments to address a problem. & “Propose a multimodal tree-search method.” \\
Evaluate & Assessing results using structured criteria or benchmarks. & “Compare keypoint coverage across baselines.” \\

\bottomrule
\end{tabular}
}
\caption{Core terminology used in our benchmark, grouped into inspiration, its subtypes, and research task phases. This table standardizes interpretation of key concepts throughout the paper.}
\label{tab:core-terms}
\end{table*}

      

\paragraph{Data Collection.}
To support the construction of research tasks grounded in authentic scholarly practice, we curated a diverse corpus of over 200 academic seminar videos spanning 12 disciplines, contributed by PhD-level researchers and sourced from publicly accessible academic seminar recordings spanning multiple disciplines. Each video is knowledge-dense and typically lasts around or over 1 hour, and the disciplinary distribution of this corpus is illustrated in Figure~\ref{fig:main}. Seminar recordings preserve the full contextual flow of expert discourse, encompassing how researchers synthesize prior knowledge, design new approaches, and evaluate outcomes. In this way, they offer a rich context for task generation. Compared to static corpora such as Wikipedia or scientific articles, seminar data captures dynamic and authentic interactions among scholars, reflecting the iterative and evolving nature of real-world research. 

As a first step in processing the raw seminar videos, we extract the audio and convert it into textual transcripts with automatic speech recognition. The resulting transcripts retain the full semantic content of the original recordings while remaining absent from existing LLM pretraining corpora, thereby reducing the risk of data contamination and ensuring the integrity of task construction.

\paragraph{Inspiration Extraction.}
Based on seminar transcripts, \textbf{Inspira Agent} automatically extracts \textit{inspirations} (as illustrated in Table \ref{tab:core-terms}) from seminar transcripts, transforming unstructured expert discourse into structured units suitable for downstream research task construction. To identify academically valuable content, the agent evaluates candidate segments along four dimensions: Novelty, Explorability, Challenge, and Verifiability. Each selected inspiration must satisfy at least two of these criteria. This multi-dimensional filtering process enables the agent to effectively suppress irrelevant or redundant material, reorganize latent research signals, and produce outputs with clearer logical structure and sharper thematic focus, thereby enhancing their suitability for subsequent task generation. In addition, the agent categorizes each item based on its informational focus into one of four types: \textit{Limitation}, \textit{Methodology}, \textit{transdisciplinarity}, \textit{Hypothesis}, as illustrated in Table \ref{tab:core-terms}, representing testable claims that can be empirically verified.

\paragraph{Task Generation and Filtering.}
Building on the structured inspirations extracted from seminar transcripts, we deploy \textbf{TaskWeaver Agent} that aggregates and reorganizes content across multiple inspirations to synthesize a focused set of concrete research tasks distributed across three key phases—\textit{Synthesize}, \textit{Design}, and \textit{Evaluate}, as illustrated in Figure~\ref{fig:main}. These tasks are constructed by identifying the core problem focus or methodological cues within the inspirations and are paired with clearly defined, executable goals. This structured synthesis process enables the scalable construction of diverse, high-quality DeepResearch tasks aligned with the demands of real-world scientific inquiry \cite{yu2022structure}.

To rank the quality of research tasks, we adopt \textbf{RankEval Agent} based on the Elo rating system~\cite{glickman1995comprehensive}. Each task is initialized with a rating score of 1200. In each round, we randomly sample disjoint pairs of tasks and compare them based on evaluation criteria such as originality, clarity, and scientific relevance.
Given a pair of tasks $t_a$ and $t_b$ with current Elo scores $r_a$ and $r_b$, we first compute the expected winning probabilities using:
\begin{equation}
e_a = \frac{1}{1 + 10^{(r_b - r_a)/400}}, 
\end{equation}
where $e_b = 1 - e_a$.
An evaluator determines which task is preferred, along with a confidence score $C \in [0.5, 1.0]$. Based on this, we assign soft outcomes:
\begin{equation}
s_a = C, \quad s_b = 1 - C
\end{equation}
We then update the Elo scores using the following update rule:
\begin{equation}
r_a' = r_a + K \cdot (s_a - e_a), \quad r_b' = r_b + K \cdot (s_b - e_b)
\end{equation}
where $K$ is a tunable constant controlling the update magnitude, set to $K=32$ in our implementation. This procedure is repeated over $R$ rounds of comparisons (e.g., $R=2$), allowing the scores to stabilize. After all rounds, we select the top $K$ tasks with the highest Elo scores as the final outputs.


\section{Evaluation Methodology}
To comprehensively assess the capabilities of deep research agents in research-oriented tasks, we propose a hybrid evaluation framework that integrates both objective and subjective dimensions of performance. Traditional benchmarks often focus narrowly on surface-level accuracy or retrieval metrics, failing to capture the nuanced reasoning, creativity, and methodological rigor required for real-world research. In contrast, our framework disentangles these facets by combining (1) Keypoint-Aligned Evaluation (KAE) to measure factual correctness and grounding against reference materials, and (2) Adaptively-generated Checklist Evaluation (ACE) to score open-ended outputs via fine-grained, model-adaptive rubrics. This dual approach enables multi-perspective assessment across all stages of the research workflow, from literature synthesis to hypothesis generation and empirical validation, offering a more faithful estimate of models’ deep research competence.

\paragraph{Keypoint-Aligned Evaluation.}
To evaluate the factual adequacy of model-generated research reports in a reference-grounded and scalable manner, we propose a structured KAE pipeline.

Let $R$ denote a model-generated report, and let $\mathit{URL}(R)$ represent the set of all cited URLs in $R$. For each URL $u \in \mathit{URL}(R)$, we retrieve the underlying webpage content and extract its factual keypoints using a keypoint extraction function $\mathit{Extract}(u)$:
\begin{equation}
K_u = \mathit{Extract}(u)
\end{equation}
We then aggregate the keypoints from all cited sources into a unified, de-duplicated list of keypoints, which we term the Unified Evidence Keypoints (UEK):
\begin{equation}
\text{UEK} = \mathit{Dedup} \left( \bigcup_{u \in \mathit{URL}(R)} K_u \right)
\end{equation}
Given this set of reference keypoints, we evaluate the report $R$ along three dimensions:

\textbf{(1) Keypoint Supported Rate (KSR):} the proportion of keypoints from UEK that are explicitly covered or supported in the report:
\begin{equation}
\text{KSR}(R) = \frac{|\mathit{Supported}(R, \text{UEK})|}{|\text{UEK}|}
\end{equation}

\textbf{(2) Keypoint Conflict Rate (KCR):} the proportion of keypoints from UEK that are contradicted by content in the report:
\begin{equation}
\text{KCR}(R) = \frac{|\mathit{conflict}(R, \text{UEK})|}{|\text{UEK}|}
\end{equation}

\textbf{(3) Keypoint Omission Rate (KOR):} the proportion of keypoints from UEK that are omitted by content in the report:
\begin{equation}
\text{KCR}(R) = \frac{|\mathit{Omitted}(R, \text{UEK})|}{|\text{UEK}|}
\end{equation}

Ideally, a high-quality research report should achieve a high KSR (indicating comprehensive factual coverage) and a low KCR and KOR (indicating consistency with evidence). These metrics enable interpretable, reference-grounded evaluation of factual alignment.

\newcommand{\stagecell}[2]{%
  \begin{tabular}{@{}ll@{}}#1 & #2 \end{tabular}} 
  
\begin{table*}[!t]
\centering
\small
\setlength{\tabcolsep}{5pt}
\renewcommand{\arraystretch}{1.2}

\begin{tabular}{p{5cm}||*{3}{c|} c| c| c}
\toprule
\multirow{2}{*}{\textbf{Model}} & 
\multicolumn{3}{c|}{\textbf{KAE}} & 
\multirow{2}{*}{\textbf{ACE}} & 
\multirow{2}{*}{\textbf{Avg. Token (k)}} & 
\multirow{2}{*}{\textbf{Avg. references}} \\
\cline {2-4}
& \textbf{KSR} & \textbf{KCR} & \textbf{KOR} & &\\

\midrule
gpt-4o-search-preview     & \stagecell{50.0}{\textbf{85.0}} & \stagecell{8.9}{5.0} & \stagecell{41.1}{10.0} & \stagecell{2.41}{2.00} & \stagecell{1.21}{2.85} & \stagecell{4.24}{3.49} \\
gpt-4o-mini-search-preview   & \stagecell{\underline{78.7}}{55.6} & \stagecell{8.5}{16.7} & \stagecell{\underline{12.8}}{27.8} & \stagecell{2.23}{2.05} & \stagecell{\textbf{1.07}}{\underline{2.23}} & \stagecell{3.83}{2.07} \\
gpt-4.1-mini  w/search           & \stagecell{62.5}{76.5} & \stagecell{10.9}{\textbf{5.9}} & \stagecell{26.6}{17.6} & \stagecell{2.21}{1.87} & \stagecell{\underline{1.10}}{\textbf{2.02}}  & \stagecell{4.75}{2.39} \\
gpt-4.1 w/search & \stagecell{77.8}{60.6} & \stagecell{\textbf{2.8}}{\underline{6.1}}& \stagecell{19.4}{33.3} & \stagecell{2.43}{2.22} & \stagecell{1.20}{2.43} & \stagecell{3.51}{2.44} \\
o4-mini-deepresearch  & \stagecell{77.2}{75.8} & \stagecell{4.3}{18.2} & \stagecell{18.5}{\underline{6.1}} & \stagecell{\textbf{4.03}}{\underline{3.88}} & \stagecell{5.59}{12.5} & \stagecell{\textbf{29.66}}{\textbf{37.27}} \\
gemini-2.5-pro w/search & \stagecell{65.1}{76} & \stagecell{14.3}{12} & \stagecell{20.6}{12} & \stagecell{2.97}{\textbf{4.03}} & \stagecell{4.29}{9.14} & \stagecell{23.86}{21.39} \\
gemini-2.5-flash w/search& \stagecell{\underline{78.7}}{\underline{80}} & \stagecell{\underline{3.4}}{16} & \stagecell{18}{\textbf{4}} & \stagecell{\underline{3.81}}{3.58} & \stagecell{64.09}{19.78} & \stagecell{\underline{29.54}}{\underline{28.07}} \\
grok-4 w/search     & \stagecell{\textbf{83.3}}{50} & \stagecell{7.5}{13.8} & \stagecell{\textbf{9.2}}{36.2} & \stagecell{2.97}{2.97} & \stagecell{3.16}{6.60} & \stagecell{20.59}{19.95} \\
\bottomrule
\end{tabular}
\caption{Evaluation metrics across models. The model release dates are omitted for brevity. Each column reports two values, with the left representing the evaluation results on the English task and the right on the Chinese task.}
\label{tab:model-metrics}
\end{table*}

\paragraph{Adaptively-generated Checklist Evaluation.}
To address the challenges of evaluating open-ended research tasks that lack fixed reference answers, we introduce Adaptively-generated Checklist Evaluation (ACE), a two-stage evaluation protocol that leverages the analytical capabilities of large language models (LLMs) while mitigating common sources of bias and inconsistency.

In the first stage, we use a high-capacity LLM (e.g., GPT-4o) to perform meta-analysis over the task prompt, generating a customized checklist of evaluation criteria tailored to the query. Each checklist item corresponds to a critical evaluation dimension, such as factual correctness, methodological soundness, formatting, or reasoning clarity, and is assigned a normalized weight to reflect its relative importance. This step serves to concretize abstract judgment into discrete, model-understandable subgoals.

In the second stage, a separate LLM is tasked with scoring the model-generated response against the checklist. For each item, the evaluator model independently assesses whether the response satisfies the criterion and assigns a local score. These individual scores are then aggregated via a weighted average to produce a final task-level rating. By decoupling checklist generation from scoring, ACE reduces evaluation bias, especially those arising from the evaluator's limited comprehension or heuristic shortcuts.

ACE addresses key limitations of existing evaluation paradigms. Human evaluation, while often considered the gold standard, suffers from subjectivity, inter-annotator inconsistency, and high cost. LLM-as-a-judge methods, especially when using smaller models, struggle with complex query understanding, detailed analytical reasoning, and accurate interpretation. Furthermore, rubric-based methods either rely on static reference answers, which are unsuitable for open-ended tasks, or require hand-crafted criteria that are difficult to scale and generalize. In contrast, ACE provides a flexible, scalable, and more reliable alternative for nuanced research task evaluation.

\section{Experiments}

\begin{figure*}[!t]
\centering
\includegraphics[width=0.95\linewidth]{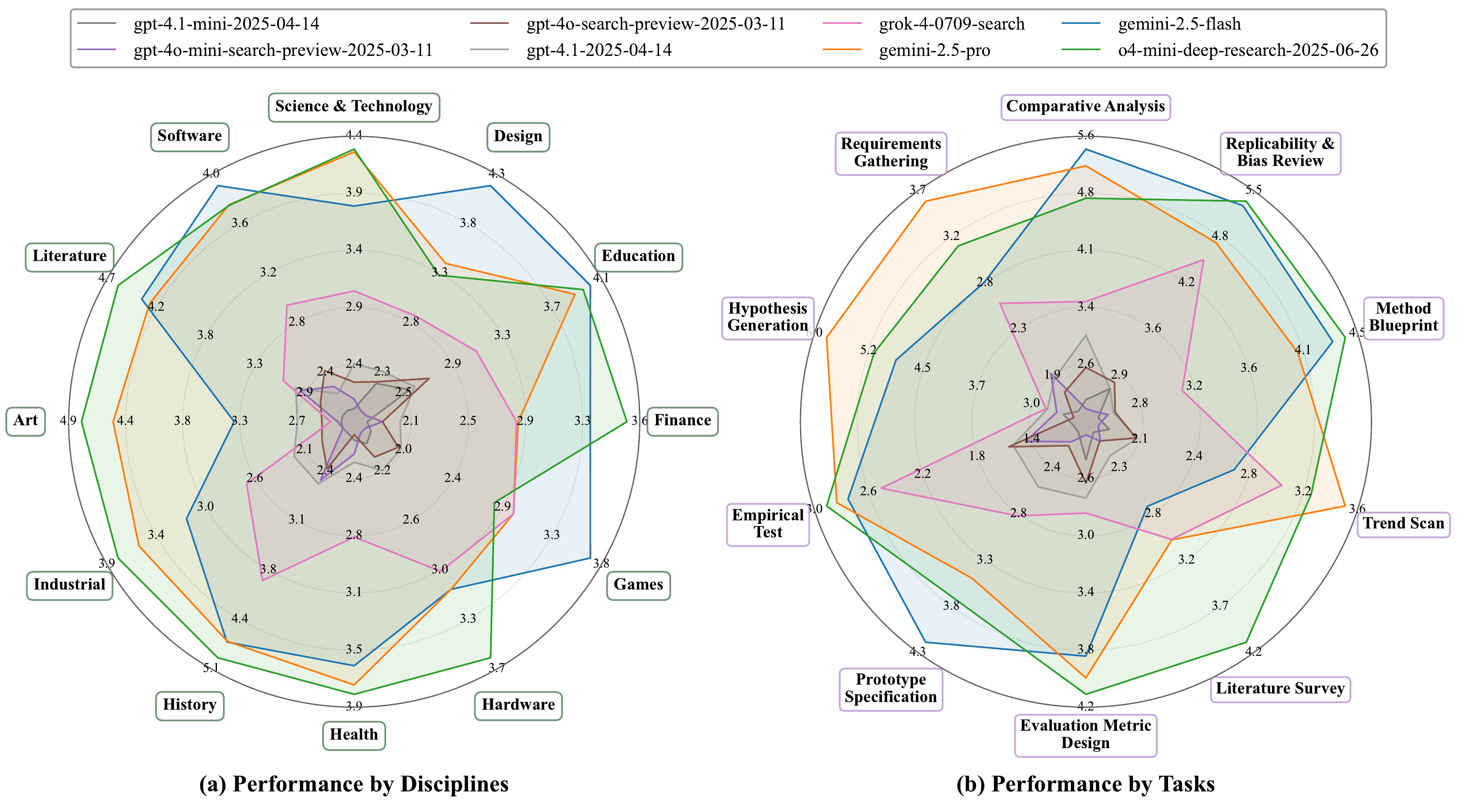}
\caption{Comparison of current mainstream models on the DeepResearch Arena benchmark. (a) Performance across 12 research disciplines (\textit{e.g.}, Science \& Technology, Art, Finance). (b) Performance across 10 research task types (\textit{e.g.}, Hypothesis Generation, Method Blueprint, Evaluation Metric Design), highlighting task-specific capabilities.} \label{radar_chart}
\end{figure*}

\begin{figure}[!t]
    \centering
    \includegraphics[width=0.95\linewidth]{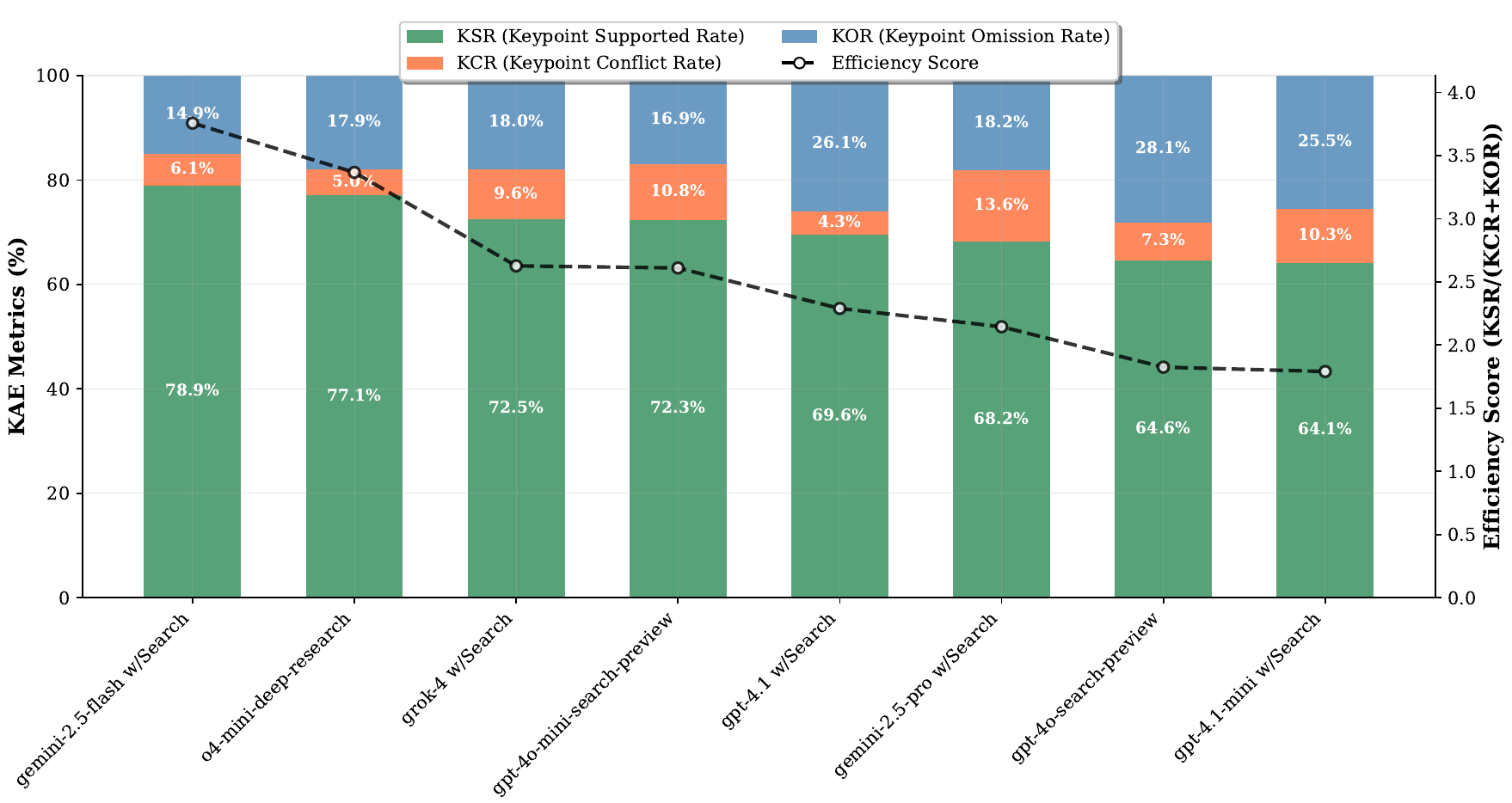}
    \caption{Comparison of DeepResearch agents in terms of Keypoint-Aligned Evaluation (KAE) metrics and efficiency.}
    \label{fig:metrics_composition}
\end{figure}

\paragraph{Implementation Details.}
Our MAHTG system comprises several specialized agents, each responsible for a distinct stage in transforming raw academic seminars into structured research tasks and evaluations.

\noindent \textbf{Model Selection Rationale.}  
We adopt a \textit{heterogeneous model configuration} across the MAHTG system, guided by three principles: (1) \textit{capability-task alignment}, assigning models suited to each agent’s functional role; (2) \textit{cost-effectiveness and scalability}, ensuring efficiency over large-scale data; and (3) \textit{robustness through model diversity}, mitigating systemic bias. Large models like \textit{claude-sonnet-4-20250514} are used for structured reasoning and code-like outputs, while lightweight ones like \textit{gpt-4o-mini} support tasks requiring relative preference. The Inspira Agent adopts \textit{claude-sonnet-4-20250514} for its strong long-context handling and structured generation. The same model powers the TaskWeaver Agent to ensure schema consistency in transforming inspirations into structured tasks. For efficient pairwise evaluation, the RankEval Agent uses \textit{gpt-4o-mini}, balancing accuracy and cost under the ELO-based framework. To reduce costs, we selected the top 100 highest-scoring samples from the full dataset for evaluation.
The choices for LLM align with human-like action and are empirically validated (see appendix).

We use \textit{gemini-2.5-flash} as a unified evaluator for both factual and subjective scoring, leveraging its strong instruction-following and long-context reasoning. In KAE, it extracts key factual statements from sources retrieved via the Jina AI API and determines whether each is supported, contradicted, or omitted. In ACE, it generates detailed, task-specific checklists and conducts criterion-based evaluation. This setup ensures consistency across evaluation stages while maintaining precision, scalability, and interpretability.

\paragraph{Evaluated Models.}
We evaluate a diverse suite of large language models covering both frontier-level deep research agents and models augmented with real-time retrieval capabilities. Specifically, we include \textit{gpt-4o-search-preview-2025-03-11}, \textit{gpt-4o-mini-search-preview-2025-03-11}, \textit{gpt-4.1-2025-04-14 w/search}, \textit{gpt-4.1-mini-2025-04-14 w/search}, \textit{o4-mini-deepresearch-2025-06-26}, \textit{gemini-2.5-pro w/search}, \textit{gemini-2.5-flash w/search}, and \textit{grok-4-0709 w/search}. When referring to these models in the future, abbreviations will be used, ignoring with search and time versions.

\paragraph{Overall Performance.}
The table~\ref{tab:model-metrics}reveals clear differences in both ACE and KCE across models. The best ACE performance is achieved by \textit{gpt-o4-mini-deep-research}, which combines the highest ACE score of 4.03 with strong KAE metrics, demonstrating accurate, well-structured, and comprehensive outputs. \textit{GPT-4.1} excels in factual precision but falls short in subjective quality, with the lowest KCR. It minimizes factual errors, yet its lower ACE scores suggest limited coherence and depth.
\textit{Gemini-2.5-flash} also performs strongly, with relatively high factual coverage and low contradiction and omission, though it uses significantly more tokens than any other model, indicating a trade-off between thoroughness and efficiency. 
In contrast, \textit{gpt-4o-search-preview}and \textit{gpt-4o-mini-search-preview} use far fewer tokens but do not perform so well in both evaluation dimensions, suggesting limited ability to handle complex research tasks. 
\textit{grok-4} demonstrates the strongest factual grounding on English tasks (KSR 83.3), but its performance drops sharply in Chinese, with significantly lower coverage and higher omission. This highlights its limited multilingual generalization despite strong English capabilities.
Overall, the results reflect varying model strengths, with some excelling in precision and others in depth or efficiency.

\paragraph{Performance on Different Tasks.}
As shown in Figure~\ref{radar_chart}, the ACE-based subjective evaluation reveals substantial differences in how models perform across various research task types. 
Models like \textit{gpt-o4-mini-deepresearch} and \textit{gemini-2.5-flash} demonstrate consistently strong performance across nearly all tasks, especially excelling in complex and high-level tasks such as hypothesis generation, evaluation metric design, and methodological planning. \textit{Gemini-2.5-pro} also shows well-rounded capabilities, performing reliably in tasks that require comparative analysis and methodological reasoning. The \textit{gpt-4o} family, particularly the mini version, performs poorly across most task types, struggling especially with tasks that require multi-step logic and structured outputs. These differences highlight each model’s unique strengths and limitations, underscoring the importance of task-specific evaluation in assessing deep research competence.

Models also show clear differences in task performance under the KAE as shown in Figure~\ref{fig:metrics_composition}. \textit{Gemini-2.5-flash} and \textit{gpt-o4-mini-deepresearch} achieve the strongest overall results, with high keypoint coverage and low conflict and omission rates, leading to the highest efficiency scores. 
In contrast, \textit{gemini-2.5-pro}, \textit{gpt-4o-search-preview}, and \textit{gpt-4.1-mini} struggle with higher conflict and omission rates, resulting in the lowest efficiency and limited reliability for fact-intensive generation. Overall, the results highlight substantial differences in how models handle task complexity and factual alignment, underscoring the value of KAE for fine-grained evaluation of research capabilities.

The experiment for effectiveness of our benchmark to prevent data leakage is detailed in the appendix.

\section{Conclusion}
We present the \textit{\textbf{DeepResearch Arena}}, a novel benchmark for evaluating the deep research capabilities of large language models in realistic, open-ended settings. Grounded in cognitive theories and authentic seminar discourse, DeepResearch Arena captures the contextual complexity and methodological ambiguity of real-world research. It systematically assesses LLM-based agents across three essential stages, through a curated corpus of multidisciplinary seminars, a hierarchical task generation pipeline, and a hybrid evaluation protocol measuring both factual grounding and higher-order reasoning. By bridging the gap between retrieval-centric agent design and cognitively demanding research tasks, it offers a rigorous, theory-aligned foundation for advancing next-generation research assistants.

\bibliography{aaai2026}

@misc{google2025deepresearch,
  author       = {Google},
  title        = {Deep Research is now available on Gemini 2.5 Pro Experimental},
  howpublished = {Gemini Blog (online)},
  month        = apr,
  year         = {2025},
  note         = {Gemini Advanced subscribers can use Deep Research powered by Gemini 2.5 Pro Experimental},
  url          = {https://blog.google/products/gemini/deep-research-gemini-2-5-pro-experimental/}
}

@misc{openai2025deepresearch,
  author       = {OpenAI},
  title        = {Introducing Deep Research},
  howpublished = {\url{https://cdn.openai.com/API/docs/deep_research_blog.pdf?utm_source=chatgpt.com}},
  month        = feb,
  year         = {2025},
  note         = {Accessed July 30, 2025},
}

@misc{huang2025deepresearchagentssystematic,
      title={Deep Research Agents: A Systematic Examination And Roadmap}, 
      author={Yuxuan Huang and Yihang Chen and Haozheng Zhang and Kang Li and Meng Fang and Linyi Yang and Xiaoguang Li and Lifeng Shang and Songcen Xu and Jianye Hao and Kun Shao and Jun Wang},
      year={2025},
      eprint={2506.18096},
      archivePrefix={arXiv},
      primaryClass={cs.AI},
      url={https://arxiv.org/abs/2506.18096}, 
}

@misc{xu2025comprehensivesurveydeepresearch,
      title={A Comprehensive Survey of Deep Research: Systems, Methodologies, and Applications}, 
      author={Renjun Xu and Jingwen Peng},
      year={2025},
      eprint={2506.12594},
      archivePrefix={arXiv},
      primaryClass={cs.AI},
      url={https://arxiv.org/abs/2506.12594}, 
}

@article{jonassen1997isp,
  title = {Instructional design models for well-structured and ill-structured problem-solving learning outcomes},
  author = {Jonassen, David H.},
  journal = {Educational Technology Research and Development},
  volume = {45},
  number = {1},
  pages = {65--94},
  year = {1997},
  publisher = {Springer},
  doi = {10.1007/BF02299613}
}

@misc{xai2025grok3,
  author       = {{xAI}},
  title        = {Grok 3},
  year         = {2025},
  howpublished = {\url{https://x.ai/news/grok-3}},
  note         = {Accessed: 2025-07-30}
}

@inproceedings{yang2018hotpotqa,
  title={{HotpotQA}: A Dataset for Diverse, Explainable Multi-hop Question Answering},
  author={Yang, Zhilin and Qi, Peng and Zhang, Saizheng and Bengio, Yoshua and Cohen, William W. and Salakhutdinov, Ruslan and Manning, Christopher D.},
  booktitle={Conference on Empirical Methods in Natural Language Processing ({EMNLP})},
  year={2018}
}

@article{geva-etal-2021-aristotle,
    title = "Did Aristotle Use a Laptop? A Question Answering Benchmark with Implicit Reasoning Strategies",
    author = "Geva, Mor  and
      Khashabi, Daniel  and
      Segal, Elad  and
      Khot, Tushar  and
      Roth, Dan  and
      Berant, Jonathan",
    editor = "Roark, Brian  and
      Nenkova, Ani",
    journal = "Transactions of the Association for Computational Linguistics",
    volume = "9",
    year = "2021",
    address = "Cambridge, MA",
    publisher = "MIT Press",
    url = "https://aclanthology.org/2021.tacl-1.21/",
    doi = "10.1162/tacl_a_00370",
    pages = "346--361"
}

@article{Ott_2023,
   title={ThoughtSource: A central hub for large language model reasoning data},
   volume={10},
   ISSN={2052-4463},
   url={http://dx.doi.org/10.1038/s41597-023-02433-3},
   DOI={10.1038/s41597-023-02433-3},
   number={1},
   journal={Scientific Data},
   publisher={Springer Science and Business Media LLC},
   author={Ott, Simon and Hebenstreit, Konstantin and Liévin, Valentin and Hother, Christoffer Egeberg and Moradi, Milad and Mayrhauser, Maximilian and Praas, Robert and Winther, Ole and Samwald, Matthias},
   year={2023},
   month=aug }

@misc{trivedi2022musiquemultihopquestionssinglehop,
      title={MuSiQue: Multihop Questions via Single-hop Question Composition}, 
      author={Harsh Trivedi and Niranjan Balasubramanian and Tushar Khot and Ashish Sabharwal},
      year={2022},
      eprint={2108.00573},
      archivePrefix={arXiv},
      primaryClass={cs.CL},
      url={https://arxiv.org/abs/2108.00573}, 
}

@inproceedings{rein2024gpqa,
      title={{GPQA}: A Graduate-Level Google-Proof Q\&A Benchmark},
      author={David Rein and Betty Li Hou and Asa Cooper Stickland and Jackson Petty and Richard Yuanzhe Pang and Julien Dirani and Julian Michael and Samuel R. Bowman},
      booktitle={First Conference on Language Modeling},
      year={2024},
      url={https://openreview.net/forum?id=Ti67584b98}
}

@misc{wu2025agenticreasoningstreamlinedframework,
      title={Agentic Reasoning: A Streamlined Framework for Enhancing LLM Reasoning with Agentic Tools}, 
      author={Junde Wu and Jiayuan Zhu and Yuyuan Liu and Min Xu and Yueming Jin},
      year={2025},
      eprint={2502.04644},
      archivePrefix={arXiv},
      primaryClass={cs.AI},
      url={https://arxiv.org/abs/2502.04644}, 
}

@misc{phan2025humanitysexam,
      title={Humanity's Last Exam}, 
      author={Long Phan and Alice Gatti and Ziwen Han and Nathaniel Li and Josephina Hu and Hugh Zhang and et al.},
      year={2025},
      eprint={2501.14249},
      archivePrefix={arXiv},
      primaryClass={cs.LG},
      url={https://arxiv.org/abs/2501.14249}, 
}

@misc{lu2022learnexplainmultimodalreasoning,
      title={Learn to Explain: Multimodal Reasoning via Thought Chains for Science Question Answering}, 
      author={Pan Lu and Swaroop Mishra and Tony Xia and Liang Qiu and Kai-Wei Chang and Song-Chun Zhu and Oyvind Tafjord and Peter Clark and Ashwin Kalyan},
      year={2022},
      eprint={2209.09513},
      archivePrefix={arXiv},
      primaryClass={cs.CL},
      url={https://arxiv.org/abs/2209.09513}, 
}

@article{clark2018think,
  author    = {Peter Clark and Isaac Cowhey and Oren Etzioni and Tushar Khot and
               Ashish Sabharwal and Carissa Schoenick and Oyvind Tafjord},
  title     = {Think You Have Solved Question Answering? Try {ARC}, the {AI2}
               Reasoning Challenge},
  journal   = {arXiv preprint arXiv:1803.05457},
  year      = {2018}
}

@misc{glazer2024frontiermathbenchmarkevaluatingadvanced,
      title={FrontierMath: A Benchmark for Evaluating Advanced Mathematical Reasoning in AI}, 
      author={Elliot Glazer and Ege Erdil and Tamay Besiroglu and Diego Chicharro and Evan Chen and Alex Gunning and Caroline Falkman Olsson and Jean-Stanislas Denain and Anson Ho and Emily de Oliveira Santos and Olli Järviniemi and Matthew Barnett and Robert Sandler and Matej Vrzala and Jaime Sevilla and Qiuyu Ren and Elizabeth Pratt and Lionel Levine and Grant Barkley and Natalie Stewart and Bogdan Grechuk and Tetiana Grechuk and Shreepranav Varma Enugandla and Mark Wildon},
      year={2024},
      eprint={2411.04872},
      archivePrefix={arXiv},
      primaryClass={cs.AI},
      url={https://arxiv.org/abs/2411.04872}, 
}

@misc{laurent2024labbenchmeasuringcapabilitieslanguage,
      title={LAB-Bench: Measuring Capabilities of Language Models for Biology Research}, 
      author={Jon M. Laurent and Joseph D. Janizek and Michael Ruzo and Michaela M. Hinks and Michael J. Hammerling and Siddharth Narayanan and Manvitha Ponnapati and Andrew D. White and Samuel G. Rodriques},
      year={2024},
      eprint={2407.10362},
      archivePrefix={arXiv},
      primaryClass={cs.AI},
      url={https://arxiv.org/abs/2407.10362}, 
}

@article{du2025deepresearch,
  author    = {Mingxuan Du and Benfeng Xu and Chiwei Zhu and Xiaorui Wang and Zhendong Mao},
  title     = {DeepResearch Bench: A Comprehensive Benchmark for Deep Research Agents},
  journal   = {arXiv preprint},
  year      = {2025},
}

@misc{li2025adaptivetooluselarge,
      title={Adaptive Tool Use in Large Language Models with Meta-Cognition Trigger}, 
      author={Wenjun Li and Dexun Li and Kuicai Dong and Cong Zhang and Hao Zhang and Weiwen Liu and Yasheng Wang and Ruiming Tang and Yong Liu},
      year={2025},
      eprint={2502.12961},
      archivePrefix={arXiv},
      primaryClass={cs.AI},
      url={https://arxiv.org/abs/2502.12961}, 
}

@article{Qu_2025,
   title={Tool learning with large language models: a survey},
   volume={19},
   ISSN={2095-2236},
   url={http://dx.doi.org/10.1007/s11704-024-40678-2},
   DOI={10.1007/s11704-024-40678-2},
   number={8},
   journal={Frontiers of Computer Science},
   publisher={Springer Science and Business Media LLC},
   author={Qu, Changle and Dai, Sunhao and Wei, Xiaochi and Cai, Hengyi and Wang, Shuaiqiang and Yin, Dawei and Xu, Jun and Wen, Ji-rong},
   year={2025},
   month=jan }

@misc{tang2023toolalpacageneralizedtoollearning,
      title={ToolAlpaca: Generalized Tool Learning for Language Models with 3000 Simulated Cases}, 
      author={Qiaoyu Tang and Ziliang Deng and Hongyu Lin and Xianpei Han and Qiao Liang and Boxi Cao and Le Sun},
      year={2023},
      eprint={2306.05301},
      archivePrefix={arXiv},
      primaryClass={cs.CL},
      url={https://arxiv.org/abs/2306.05301}, 
}

@article{baek2024researchagent,
  title={Researchagent: Iterative research idea generation over scientific literature with large language models},
  author={Baek, Jinheon and Jauhar, Sujay Kumar and Cucerzan, Silviu and Hwang, Sung Ju},
  journal={arXiv preprint arXiv:2404.07738},
  year={2024}
}

@article{li2024chain,
  title={Chain of ideas: Revolutionizing research via novel idea development with llm agents},
  author={Li, Long and Xu, Weiwen and Guo, Jiayan and Zhao, Ruochen and Li, Xingxuan and Yuan, Yuqian and Zhang, Boqiang and Jiang, Yuming and Xin, Yifei and Dang, Ronghao and others},
  journal={arXiv preprint arXiv:2410.13185},
  year={2024}
}

@misc{schmidgall2025agentlaboratoryusingllm,
      title={Agent Laboratory: Using LLM Agents as Research Assistants}, 
      author={Samuel Schmidgall and Yusheng Su and Ze Wang and Ximeng Sun and Jialian Wu and Xiaodong Yu and Jiang Liu and Michael Moor and Zicheng Liu and Emad Barsoum},
      year={2025},
      eprint={2501.04227},
      archivePrefix={arXiv},
      primaryClass={cs.HC},
      url={https://arxiv.org/abs/2501.04227}, 
}

@book{einstein1938evolution,
  title     = {The Evolution of Physics},
  author    = {Einstein, Albert and Infeld, Leopold},
  year      = {1938},
  publisher = {Simon and Schuster},
  pages     = {92}
}

@article{zhou2025academicbrowse,
  title={AcademicBrowse: Benchmarking Academic Browse Ability of LLMs},
  author={Zhou, Junting and Li, Wang and Liao, Yiyan and Zhang, Nengyuan and Qi, Tingjia Miaoand Zhihui and Wu, Yuhan and Yang, Tong},
  journal={arXiv preprint arXiv:2506.13784},
  year={2025}
}

@article{wei2025browsecomp,
  title={Browsecomp: A simple yet challenging benchmark for browsing agents},
  author={Wei, Jason and Sun, Zhiqing and Papay, Spencer and McKinney, Scott and Han, Jeffrey and Fulford, Isa and Chung, Hyung Won and Passos, Alex Tachard and Fedus, William and Glaese, Amelia},
  journal={arXiv preprint arXiv:2504.12516},
  year={2025}
}

@article{liu2025researchbench,
  title={Researchbench: Benchmarking llms in scientific discovery via inspiration-based task decomposition},
  author={Liu, Yujie and Yang, Zonglin and Xie, Tong and Ni, Jinjie and Gao, Ben and Li, Yuqiang and Tang, Shixiang and Ouyang, Wanli and Cambria, Erik and Zhou, Dongzhan},
  journal={arXiv preprint arXiv:2503.21248},
  year={2025}
}

@article{ruan2025expertlongbench,
  title={ExpertLongBench: Benchmarking Language Models on Expert-Level Long-Form Generation Tasks with Structured Checklists},
  author={Ruan, Jie and Nair, Inderjeet and Cao, Shuyang and Liu, Amy and Munir, Sheza and Pollens-Dempsey, Micah and Chiang, Tiffany and Kates, Lucy and David, Nicholas and Chen, Sihan and others},
  journal={arXiv preprint arXiv:2506.01241},
  year={2025}
}

@article{glickman1995comprehensive,
  title={A comprehensive guide to chess ratings},
  author={Glickman, Mark E},
  year={1995},
  journal={American Chess Journal},
 volume={3},
issue={1}
}

@misc{xu2024benchmarkingbenchmarkleakagelarge,
      title={Benchmarking Benchmark Leakage in Large Language Models}, 
      author={Ruijie Xu and Zengzhi Wang and Run-Ze Fan and Pengfei Liu},
      year={2024},
      eprint={2404.18824},
      archivePrefix={arXiv},
      primaryClass={cs.CL},
      url={https://arxiv.org/abs/2404.18824}, 
}

@misc{zhou2025lessleakbenchinvestigationdataleakage,
      title={LessLeak-Bench: A First Investigation of Data Leakage in LLMs Across 83 Software Engineering Benchmarks}, 
      author={Xin Zhou and Martin Weyssow and Ratnadira Widyasari and Ting Zhang and Junda He and Yunbo Lyu and Jianming Chang and Beiqi Zhang and Dan Huang and David Lo},
      year={2025},
      eprint={2502.06215},
      archivePrefix={arXiv},
      primaryClass={cs.SE},
      url={https://arxiv.org/abs/2502.06215}, 
}

@article{ozturk2016comparative,
  title={A comparative study of SMILES-based compound similarity functions for drug-target interaction prediction},
  author={{\"O}zt{\"u}rk, Hakime and Ozkirimli, Elif and {\"O}zg{\"u}r, Arzucan},
  journal={BMC bioinformatics},
  volume={17},
  number={1},
  pages={128},
  year={2016},
  publisher={Springer}
}

@Article{info11090421,
AUTHOR = {Wang, Jiapeng and Dong, Yihong},
TITLE = {Measurement of Text Similarity: A Survey},
JOURNAL = {Information},
VOLUME = {11},
YEAR = {2020},
NUMBER = {9},
ARTICLE-NUMBER = {421},
URL = {https://www.mdpi.com/2078-2489/11/9/421},
ISSN = {2078-2489},
DOI = {10.3390/info11090421}
}

@inproceedings{castano-etal-2016-machine,
    title = "A Machine Learning Approach to Clinical Terms Normalization",
    author = "Casta{\~n}o, Jos{\'e}  and
      Gambarte, Mar{\'i}a Laura  and
      Park, Hee Joon  and
      del Pilar Avila Williams, Maria  and
      P{\'e}rez, David  and
      Campos, Fernando  and
      Luna, Daniel  and
      Ben{\'i}tez, Sonia  and
      Berinsky, Hern{\'a}n  and
      Zanetti, Sof{\'i}a",
    editor = "Cohen, Kevin Bretonnel  and
      Demner-Fushman, Dina  and
      Ananiadou, Sophia  and
      Tsujii, Jun-ichi",
    booktitle = "Proceedings of the 15th Workshop on Biomedical Natural Language Processing",
    month = aug,
    year = "2016",
    address = "Berlin, Germany",
    publisher = "Association for Computational Linguistics",
    url = "https://aclanthology.org/W16-2901/",
    doi = "10.18653/v1/W16-2901",
    pages = "1--11"
}

@article{jindal2024survey,
  title={A Survey of Text Similarity Approaches},
  author={Jindal, Sonia and Leema, M},
  journal={Journal of Artificial Intelligence and Capsule Networks},
  year={2024},
  volume={4},
  number={1},
  pages={33--45},
  doi={10.61089/jaincn-040107},
  pmcid={PMC11977957},
  url={https://www.ncbi.nlm.nih.gov/pmc/articles/PMC11977957/}
}

@inproceedings{De_Boom_2015,
   title={Learning Semantic Similarity for Very Short Texts},
   url={http://dx.doi.org/10.1109/ICDMW.2015.86},
   DOI={10.1109/icdmw.2015.86},
   booktitle={2015 IEEE International Conference on Data Mining Workshop (ICDMW)},
   publisher={IEEE},
   author={De Boom, Cedric and Van Canneyt, Steven and Bohez, Steven and Demeester, Thomas and Dhoedt, Bart},
   year={2015},
   month=nov, pages={1229–1234} }

@article{spearman1904association,
  title={The proof and measurement of association between two things},
  author={Spearman, Charles},
  journal={The American Journal of Psychology},
  volume={15},
  number={1},
  pages={72--101},
  year={1904},
  publisher={JSTOR},
  doi={10.2307/1412159}
}

@article{pearson1895regression,
  title={Note on Regression and Inheritance in the Case of Two Parents},
  author={Pearson, Karl},
  journal={Proceedings of the Royal Society of London},
  volume={58},
  pages={240--242},
  year={1895},
  publisher={The Royal Society},
  doi={10.1098/rspl.1895.0041}
}

@article{kendall1938rank,
  title={A New Measure of Rank Correlation},
  author={Kendall, Maurice G},
  journal={Biometrika},
  volume={30},
  number={1-2},
  pages={81--93},
  year={1938},
  publisher={Oxford University Press},
  doi={10.2307/2332226}
}

@inproceedings{carlini2021extracting,
  title={Extracting training data from large language models},
  author={Carlini, Nicholas and Tramer, Florian and Wallace, Eric and Jagielski, Matthew and Herbert-Voss, Ariel and Lee, Katherine and Roberts, Adam and Brown, Tom and Song, Dawn and Erlingsson, Ulfar and others},
  booktitle={30th USENIX security symposium (USENIX Security 21)},
  pages={2633--2650},
  year={2021}
}

@article{lehman2021does,
  title={Does BERT pretrained on clinical notes reveal sensitive data?},
  author={Lehman, Eric and Jain, Sarthak and Pichotta, Karl and Goldberg, Yoav and Wallace, Byron C},
  journal={arXiv preprint arXiv:2104.07762},
  year={2021}
}

@article{yu2023thought,
  title={Thought propagation: An analogical approach to complex reasoning with large language models},
  author={Yu, Junchi and He, Ran and Ying, Rex},
  journal={arXiv preprint arXiv:2310.03965},
  year={2023}
}

@article{yu2022structure,
  title={Structure-aware conditional variational auto-encoder for constrained molecule optimization},
  author={Yu, Junchi and Xu, Tingyang and Rong, Yu and Huang, Junzhou and He, Ran},
  journal={Pattern Recognition},
  volume={126},
  pages={108581},
  year={2022},
  publisher={Elsevier}
}

\clearpage
\appendix
\setcounter{secnumdepth}{2}       
\renewcommand{\thesection}{\Alph{section}}  
\renewcommand{\thesubsection}{\thesection.\arabic{subsection}}  
\section{Data Leakage Detection}

To verify that our benchmark minimize the risk of data leakage from the pretraining corpora of LLMs, we conduct a comprehensive \textit{leakage simulation experiment}~\cite{xu2024benchmarkingbenchmarkleakagelarge, zhou2025lessleakbenchinvestigationdataleakage} across all 8 evaluated models. This procedure estimates whether any model can reproduce the withheld portion of a task when prompted with only the first half of the task description.

\subsection{Experimental Procedure}

Given a task instance \( T \), we split it into two parts at a punctuation boundary \( i^* \) closest to the midpoint:
\begin{equation}
i^* = \arg\min_{i \in \mathcal{P}} \left| i - \frac{|T|}{2} \right|, \quad \mathcal{P} = \{ j \mid T[j] \in \text{punctuation} \} 
\label{eq:punct-split}
\end{equation}

Let \( \mathcal{M}_1, \dots, \mathcal{M}_8 \) denote the 8 models evaluated in the main paper. Each model \( \mathcal{M}_k \) is queried with the prompt \( T_{\text{prefix}} \), yielding a generated continuation:
\begin{equation}
\hat{T}^{(k)}_{\text{suffix}} = \mathcal{M}_k(T_{\text{prefix}})
\label{eq:model-output}
\end{equation}
where $T_{\text{prefix}} = T[:i^*]$, $ T_{\text{suffix}} = T[i^*:]$. This formulation allows us to compare the model-generated continuation \( \hat{T}^{(k)}_{\text{suffix}} \) with the ground-truth suffix \( T_{\text{suffix}} \). If the similarity between these two sequences is unexpectedly high, even though the model only received the input prefix, it may suggest that the model has memorized or encountered the full task during pretraining, thereby posing a risk of data leakage.

\subsection{Similarity Metrics}

To assess whether \( \hat{T}^{(k)}_{\text{suffix}} \) potentially replicates the ground-truth suffix \( T_{\text{suffix}} \), we compute three types of similarity:


\paragraph{1. String Similarity.}
We compute string-level similarity between the model-generated suffix and the ground-truth suffix using the normalized \textit{Longest Common Subsequence} (LCS) metric~\cite{ozturk2016comparative,info11090421}. The similarity score for model \( \mathcal{M}_k \) is defined as:
\begin{equation}
\text{Sim}^{(k)}_{\text{string}} = \frac{2 \cdot |\text{LCS}(\hat{T}^{(k)}_{\text{suffix}}, T_{\text{suffix}})|}{|\hat{T}^{(k)}_{\text{suffix}}| + |T_{\text{suffix}}|}
\label{eq:string-similarity}
\end{equation}

\noindent
Here:
\begin{itemize}
    \item \( \mathcal{M}_k \) denotes the $k$-th evaluated model.
    \item \( T_{\text{suffix}} \) is the reference suffix (i.e., the ground-truth continuation of a given task).
    \item \( \hat{T}^{(k)}_{\text{suffix}} = \mathcal{M}_k(T_{\text{prefix}}) \) is the suffix generated by model \( \mathcal{M}_k \) when prompted with the task prefix \( T_{\text{prefix}} \).
    \item \( \text{LCS}(A, B) \) denotes the \textit{Longest Common Subsequence} between sequences \( A \) and \( B \), i.e., the longest sequence of characters that appear left-to-right (but not necessarily contiguously) in both \( A \) and \( B \).
    \item \( |\cdot| \) denotes the number of characters in a sequence.
\end{itemize}

\noindent
This normalized LCS score ranges from 0 to 1, where 1 indicates that the two sequences are identical (character order preserved), and 0 indicates no character-level overlap. The formula symmetrically normalizes the LCS length by the average length of the two sequences, ensuring robustness to differing output lengths.


\paragraph{2. TF-IDF Cosine Similarity.}
We compute lexical similarity between the generated suffix and the reference suffix using cosine similarity over their TF-IDF representations~\cite{castano-etal-2016-machine,info11090421}. The score for model \( \mathcal{M}_k \) is given by:
\begin{equation}
\text{Sim}^{(k)}_{\text{tfidf}} = \frac{\mathbf{v}^{(k)} \cdot \mathbf{v}_T}{\|\mathbf{v}^{(k)}\| \cdot \|\mathbf{v}_T\|}
\label{eq:tfidf-similarity}
\end{equation}

\noindent
Here:
\begin{itemize}
    \item \( \mathcal{M}_k \) denotes the $k$-th evaluated model.
    \item \( \hat{T}^{(k)}_{\text{suffix}} \) and \( T_{\text{suffix}} \) are the model-generated and reference suffixes, respectively.
    \item \( \mathbf{v}^{(k)} \in \mathbb{R}^d \) is the TF-IDF vector of \( \hat{T}^{(k)}_{\text{suffix}} \).
    \item \( \mathbf{v}_T \in \mathbb{R}^d \) is the TF-IDF vector of \( T_{\text{suffix}} \).
    \item \( \mathbf{v}^{(k)} \cdot \mathbf{v}_T \) denotes the dot product between the two vectors.
    \item \( \|\mathbf{v}\| \) denotes the Euclidean norm (i.e., \( \|\mathbf{v}\| = \sqrt{\sum_i v_i^2} \)) of vector \( \mathbf{v} \).
\end{itemize}

\noindent
TF-IDF vectors are computed over a fixed vocabulary, transforming each suffix into a weighted bag-of-words representation. Cosine similarity then measures the angular similarity between these two vectors, ranging from 0 (completely dissimilar) to 1 (identical in direction).


\paragraph{3. Word Overlap Ratio.}
We compute word-level lexical overlap between the generated and reference suffixes using the normalized word set intersection~\cite{jindal2024survey}. The score for model \( \mathcal{M}_k \) is defined as:
\begin{equation}
\text{Sim}^{(k)}_{\text{overlap}} = \frac{|W^{(k)} \cap W_T|}{|W_T|}
\label{eq:overlap-similarity}
\end{equation}

\noindent
Here:
\begin{itemize}
    \item \( \mathcal{M}_k \) denotes the $k$-th evaluated model.
    \item \( \hat{T}^{(k)}_{\text{suffix}} \) and \( T_{\text{suffix}} \) are the model-generated and reference suffixes, respectively.
    \item \( W^{(k)} \) is the set of unique words in \( \hat{T}^{(k)}_{\text{suffix}} \), after tokenization and lowercasing.
    \item \( W_T \) is the set of unique words in \( T_{\text{suffix}} \), processed identically.
    \item \( |A| \) denotes the cardinality (i.e., number of elements) of set \( A \).
\end{itemize}

\noindent
This metric captures the proportion of reference words that are correctly recovered in the model output, regardless of order or repetition. A higher score indicates greater lexical fidelity to the reference.


\paragraph{4. Composite Similarity.}
To obtain a unified similarity score that balances multiple aspects of textual similarity, we compute a weighted combination of the three individual metrics~\cite{De_Boom_2015}:
\begin{equation}
\text{Sim}^{(k)}_{\text{composite}} = 0.4 \cdot \text{Sim}^{(k)}_{\text{string}} + 0.4 \cdot \text{Sim}^{(k)}_{\text{tfidf}} + 0.2 \cdot \text{Sim}^{(k)}_{\text{overlap}}
\label{eq:composite-similarity}
\end{equation}

\noindent
Here:
\begin{itemize}
    \item \( \text{Sim}^{(k)}_{\text{string}} \) measures normalized character-level alignment via longest common subsequence.
    \item \( \text{Sim}^{(k)}_{\text{tfidf}} \) measures cosine similarity over TF-IDF vector representations.
    \item \( \text{Sim}^{(k)}_{\text{overlap}} \) measures word-level lexical overlap based on unique token sets.
\end{itemize}

\noindent
The weights \( (0.4, 0.4, 0.2) \) were heuristically chosen to prioritize structural and semantic similarity (string and TF-IDF), while still accounting for lexical coverage (overlap). The resulting composite score lies in the range \([0, 1]\), with higher values indicating greater alignment with the reference text.

\subsection{Leakage Criterion}

We consider a task as \textit{potentially leaked} by model \( \mathcal{M}_k \) if the composite score exceeds a threshold:
\begin{equation}
\texttt{is\_leaked}^{(k)} = \mathbb{I}\left[ \text{Sim}^{(k)}_{\text{composite}} > \tau \right], \quad \text{with } \tau = 0.7
\label{eq:leakage-decision}
\end{equation}

We set the threshold \( \tau = 0.7 \) as a conservative criterion to identify potential cases of memorization or contamination. This choice is supported by prior work on training data extraction~\cite{carlini2021extracting} and sensitive domain leakage~\cite{lehman2021does}, which report that similarity scores at or above 0.7 often correspond to memorized or verbatim training content. Such a threshold ensures high precision in detecting potential data leakage while minimizing false positives.

\subsection{Experimental Setup}

We evaluate all 8 models on the same set of 100 benchmark tasks. Each model is accessed via external API. We use a decoding temperature of 0.1, a maximum output length of 500 tokens.

For each model-task pair, we log:

\begin{itemize}
  \item Input prefix \( T_{\text{prefix}} \)
  \item Model continuation \( \hat{T}^{(k)}_{\text{suffix}} \)
  \item Reference suffix \( T_{\text{suffix}} \)
  \item Three similarity scores and a composite score
  \item Binary leakage flag
\end{itemize}

The following example illustrates a typical case where the model-generated continuation remains semantically and lexically distinct from the ground truth, thereby avoiding the leakage threshold:
\tcbset{
  colback=white,
  colframe=black,
  fonttitle=\bfseries,
  coltitle=white,
  colbacktitle=green!70!black,
  boxrule=0.8pt,
  arc=2pt,
  width=\linewidth,
  left=2mm, right=2mm, top=1mm, bottom=1mm,
  enhanced,
}

\tcbset{
  colback=white,           
  colframe=black,          
  fonttitle=\bfseries,     
  coltitle=black,          
  colbacktitle=gray!20,    
  boxrule=0.6pt,           
  arc=2pt,                 
  left=2mm, right=2mm, top=1mm, bottom=1mm,
  enhanced,
  width=\linewidth
}

\begin{tcolorbox}[title=Data Leakage Detection Example (Non-Leaked Sample)]

\textbf{Input Prefix:}

\textit{Test the effectiveness of transformer architecture visualization tools in helping software architects understand generative AI systems compared to traditional documentation}

\vspace{1ex}
\textbf{Reference Suffix:}

\textit{methods. Conduct controlled experiments measuring comprehension and decision-making quality, delivering statistical analysis results and recommendations.}

\vspace{1ex}
\textbf{Model-Generated Suffix:}

\textit{To assess the effectiveness of transformer architecture visualization tools in aiding software architects' understanding of generative AI systems compared to traditional documentation, a comprehensive...}

\vspace{1ex}
\textbf{Similarity Scores:}
\begin{itemize}[left=1em]
  \item String similarity: 15.3\%
  \item TF-IDF similarity: 9.4\%
  \item Word overlap: 14.3\%
  \item Composite similarity: 12.7\%
\end{itemize}

\vspace{0.5ex}
\textbf{Leakage Flag:} \texttt{False}

\end{tcolorbox}

\subsection{Results and Analysis}

Table~\ref{tab:leakage-stats} presents the detection summary for all 8 models.

\begin{table*}[t]
\small
\centering
\begin{tabularx}{\textwidth}{X|c|c|c|c|c|c}
\toprule
\textbf{Model} & \textbf{Leak} & \textbf{Avg. Comp.} & \textbf{Avg. StrSim} & \textbf{Avg. TFIDF} & \textbf{Avg. Overlap} & \textbf{Count} \\
\midrule
\textit{gpt-4o-search-preview}     & 0.0\% & 9.8\% & 10.4\% & 7.1\% & 14.1\% & 0 \\
\textit{gpt-4o-mini-search-preview}& 0.0\% & 9.4\% & 7.7\% & 7.2\% & 17.2\% & 0 \\
\textit{gpt-4.1}     & 0.0\% & 14.8\% & 22.4\% & 8.3\% & 12.4\% & 0 \\
\textit{gpt-4.1-mini}          & 0.0\% & 13.6\% & 17.9\% & 8.9\% & 14.4\% & 0 \\
\textit{o4-mini}     & 0.0\% & 10.5\% & 5.7\% & 9.4\% & 21.9\% & 0 \\
\textit{gemini-2.5-pro}  & 0.0\% & 13.8\% & 20.8\% & 7.5\% & 12.4\% & 0 \\
\textit{gemini-2.5-flash}& 0.0\% & 13.7\% & 22.2\% & 6.9\% & 10.1\% & 0 \\
\textit{grok-4}          & 0.0\% & 14.8\% & 22.2\% & 8.4\% & 12.7\% & 0 \\
\midrule
\textbf{Average} & 0.0\% & 12.6\% & 16.1\% & 8.0\% & 14.4\% & 0.0 \\
\bottomrule
\end{tabularx}
\caption{Average similarity scores across 100 benchmark tasks for each evaluated model. Each row reports how similar the model-generated suffix is to the ground-truth suffix, given the same task prefix. \textbf{Avg. Comp.} denotes the composite similarity score, computed as a weighted average of \textbf{Avg. StrSim} (string similarity), \textbf{Avg. TFIDF} (TF-IDF cosine similarity), and \textbf{Avg. Overlap} (token overlap rate). \textbf{Leak} shows the proportion of tasks whose composite similarity exceeds 0.7, indicating potential data leakage. \textbf{Count} reflects the number of suspected leakage cases (all zero). Averages are computed over 100 tasks per model.}
\label{tab:leakage-stats}
\end{table*}

Across all evaluated models, none of the 100 sampled tasks triggered the leakage criterion, indicating that no model exceeded the composite similarity threshold of 0.7. The average similarity scores remain consistently low across string-level, semantic, and lexical dimensions. This suggests that the generated continuations are largely dissimilar from the ground-truth suffixes and unlikely to be the result of memorization. These results provide strong evidence that our benchmark is free from pretraining contamination or data leakage.

\section{Alignment Between Automated Evaluation and Human Judgment}

\subsection{Motivation}
To ensure the reliability of our benchmark evaluations, it is essential to verify that our automated scoring metrics (KAE and ACE) align well with human judgments. This section provides a systematic analysis of their agreement with expert annotations.

\subsection{Experimental Setup}
We randomly sample a representative subset of benchmark tasks and collect human evaluations for model-generated responses. Human annotators are instructed to assess each response according to the same criteria used in our automated evaluation. Each response is rated independently by three annotators, and their scores are averaged.

\subsubsection{Metric Definitions}
We compute the following correlation coefficients between the automated scores and the averaged human scores:

\begin{itemize}
  \item \textbf{Spearman’s Rank Correlation} ($\rho$): Measures the monotonic relationship between two ranked variables ~\cite{spearman1904association}. It is computed as:
  \begin{equation}
    \rho = 1 - \frac{6 \sum_{i=1}^{n} d_i^2}{n(n^2 - 1)}
  \end{equation}
  where $d_i$ is the difference between the ranks of the $i$-th observation and $n$ is the total number of samples.

  \item \textbf{Pearson Correlation} ($r$): Measures the linear correlation between two variables $X$ and $Y$~\cite{pearson1895regression}:
  \begin{equation}
    r = \frac{\sum_{i=1}^{n} (X_i - \bar{X})(Y_i - \bar{Y})}{\sqrt{\sum_{i=1}^{n} (X_i - \bar{X})^2} \sqrt{\sum_{i=1}^{n} (Y_i - \bar{Y})^2}}
  \end{equation}
  where $\bar{X}$ and $\bar{Y}$ denote the means of $X$ and $Y$, respectively.

  \item \textbf{Kendall’s Tau} ($\tau$): Measures ordinal association by comparing concordant and discordant pairs~\cite{kendall1938rank}:
  \begin{equation}
    \tau = \frac{C - D}{\frac{1}{2} n(n - 1)}
  \end{equation}
  where $C$ and $D$ are the number of concordant and discordant pairs among all $n(n-1)/2$ possible pairs.
\end{itemize}

\subsection{Results and Analysis}

\begin{table}[h]
\centering
\begin{tabular}{lcc}
\toprule
\textbf{Metric} & \textbf{KAE vs. Human} & \textbf{ACE vs. Human} \\
\midrule
Spearman's $\rho$ & 0.84 & 0.81 \\
Pearson's $r$     & 0.79 & 0.76 \\
Kendall's $\tau$  & 0.68 & 0.65 \\
\bottomrule
\end{tabular}
\caption{Agreement between automated evaluation methods and human judgments on a representative set of benchmark tasks.}
\label{tab:human-alignment}
\end{table}

As shown in Table~\ref{tab:human-alignment}, both KAE and ACE exhibit strong correlations with human scores, particularly in terms of ranking consistency. This demonstrates that our evaluation protocol provides a faithful and efficient approximation of human preference and judgment.

\subsection{Qualitative Case Study}

To complement the quantitative analysis, we present a representative example illustrating how ACE aligns with human intuition in a realistic task setting. This case is shown in Table~\ref{tab:case-study}.

\begin{table*}[h]
\centering
\begin{tabular}{p{0.94\linewidth}}
\toprule
\textbf{Task}: \textit{Design an experimental protocol to compare the 2-approximation guarantee of greedy maximal independent set algorithms against other linear-time edge-colored clustering methods on hypergraphs. Include benchmark datasets, performance metrics for \( O(\sum \text{hyperedge sizes}) \) time complexity validation, and statistical significance testing procedures in a detailed methodology document.} \\
\midrule
\textbf{Model Output:} \\ 
\# Methodology \\
Datasets: We will evaluate on a suite of real-world hypergraphs with categorical (edge-colored) interactions that have been used as benchmarks in prior ECC research... \\
\textbf{Human Score}: 4.5/10 \quad \textbf{ACE Score}: 4.35 \\
\textbf{Comment}: Both human and ACE assign similar moderate scores, suggesting the response partially meets the task objective with a coherent but limited methodology description. \\
\bottomrule
\end{tabular}
\caption{Example of strong agreement between ACE and human judgment.}
\label{tab:case-study}
\end{table*}

\subsection{Summary}
The observed correlations and case study confirm that KAE and ACE offer scalable, interpretable, and human-aligned metrics for evaluating model performance on research tasks. They allow for reliable comparisons without incurring the cost and variability of manual annotation.

\section{Sample Checklist Generated by ACE}

We present a full example of a task-specific checklist automatically generated by a strong LLM (\textit{gemini-2.5-flash}) as part of the ACE framework. The checklist is conditioned on a challenging task in the \textit{Design} phase, requiring methodological synthesis and theoretical precision. Each criterion includes a semantic title, a detailed description, and a relative weight, summing to 1.0 across all dimensions.

\begin{figure*}[t]
\centering
\begin{tcolorbox}[title=Generated Checklist, colback=gray!5, colframe=black, width=0.95\textwidth, boxrule=0.7pt]
\small
\textbf{Task Query (Design Phase)}: \\
\emph{Design an experimental protocol to compare the 2-approximation guarantee of greedy maximal independent set algorithms against other linear-time edge-colored clustering methods on hypergraphs. Include benchmark datasets, performance metrics for $O(\sum \text{hyperedge sizes})$ time complexity validation, and statistical significance testing procedures in a detailed methodology document.}

\vspace{1em}
\textbf{Checklist Criteria:}
\begin{enumerate}[leftmargin=0.6cm, itemsep=0.7em]

\item \textbf{Methodological Rigor and Experimental Design Quality} \hfill \textbf{(Weight: 0.45)} \\
Evaluates the scientific soundness of the experimental protocol including: proper control variables, valid comparison methodology between greedy maximal independent set algorithms and edge-colored clustering methods, appropriate experimental conditions, clear hypothesis formulation, and rigorous approach to validating $O(\sum \text{hyperedge sizes})$ time complexity. Must demonstrate understanding of algorithmic analysis principles and fair comparison frameworks.

\item \textbf{Technical Accuracy and Theoretical Grounding} \hfill \textbf{(Weight: 0.30)} \\
Assesses correctness of technical concepts including: accurate understanding of 2-approximation guarantees, proper characterization of greedy maximal independent set algorithms on hypergraphs, correct complexity analysis methodology, valid performance metrics for the specified time complexity, and appropriate statistical significance testing procedures. Must demonstrate deep understanding of graph theory, approximation algorithms, and computational complexity.

\item \textbf{Completeness and Implementation Feasibility} \hfill \textbf{(Weight: 0.15)} \\
Evaluates whether the response addresses all required components: benchmark dataset specifications with hypergraph characteristics, comprehensive performance metrics beyond time complexity, detailed statistical testing procedures, practical implementation considerations, and completeness of the methodology document structure. Must provide actionable protocols that can be realistically executed.

\item \textbf{Clarity and Professional Documentation Standards} \hfill \textbf{(Weight: 0.10)} \\
Assesses the quality of presentation including: clear structure suitable for a methodology document, precise technical language, logical flow of experimental steps, appropriate level of detail for reproducibility, and professional formatting. Must be comprehensible to researchers in the field while maintaining technical precision.

\end{enumerate}

\vspace{1em}
\textbf{Checklist Metadata:}
\begin{itemize}
  \item \textbf{Generated by:} \textit{gemini-2.5-flash}
  \item \textbf{Task Type:} Method Blueprint
  \item \textbf{Task Category:} Science \& Technology
  \item \textbf{Task Difficulty:} Advanced
  \item \textbf{Video Source:} \emph{DSI Seminar Series | Algorithms and Applications of Edge-Colored Hypergraph Clustering}
\end{itemize}
\end{tcolorbox}
\end{figure*}

\clearpage

\section{Prompt Templates}
In this section, we include all the prompt templates employed during the data construction and model evaluation stages. These prompts are carefully crafted to align with the task objectives and ensure standardized interactions across models and tasks, thereby supporting transparency and reproducibility of our benchmark.

Specifically:

\begin{itemize}
  \item \textbf{INSPIRATION\_EXTRACTION\_PROMPT} \\
  Extracts categorized research inspirations (Limitation, Methodology, Transdisciplinarity, Hypothesis) from seminar transcripts, capturing authentic research motivations to seed task generation.

  \item \textbf{TASK\_GENERATOR\_PROMPT} \\
  Transforms extracted inspirations into structured DeepResearch tasks that span the full research workflow (\textit{Synthesize}, \textit{Design}, \textit{Evaluate}), grounding the benchmark in real research challenges.

  \item \textbf{RESEARCH\_TASK\_SCORING\_PROMPT} \\
  Enables head-to-head comparison of task quality, where a judge assesses competing task formulations based on clarity, specificity, feasibility, and academic value.

  \item \textbf{KEY\_POINT\_EXTRACTION\_PROMPT} \\
  Extracts key points from the content retrieved via URL associated with a research query. These points serve as targeted evidence crucial for evaluating response faithfulness.

  \item \textbf{KEY\_POINT\_RELEVANCE\_PROMPT} \\
  Evaluates whether a model-generated response appropriately reflects a specific key point, helping assess alignment with source-grounded facts or requirements.
  \item \textbf{CHECKLIST\_TEMPLATE\_PROMPT} \\
  Supports the construction of comprehensive, task-specific evaluation rubrics used to guide human or model-based scoring of open-ended responses.

  \item \textbf{SINGLE\_CRITERION\_SCORING\_PROMPT} \\
  Enables fine-grained assessment of LLM responses along a single evaluation criterion from the checklist, promoting transparency and score traceability.
\end{itemize}

\begin{figure*}[t]
\centering
\begin{tcolorbox}[title=INSPIRATION\_EXTRACTION\_PROMPT, colback=gray!5, colframe=black, width=0.95\textwidth, boxrule=0.7pt]
\small

\textbf{System Role:} \\
You are Inspiration-Extractor, an expert research assistant.

\vspace{1em}
\textbf{Goal:} \\
Read the transcript below and output a list of \emph{inspirations} — concise research leads with academic value. Each inspiration must satisfy at least \textbf{two} of the following four qualities:

\begin{itemize}
  \item \textbf{Novelty} — introduces or implies a new idea, method, or perspective.
  \item \textbf{Explorability} — offers a clear starting point for further modeling, experiments, or policy analysis.
  \item \textbf{Challenge} — exposes a limitation, bottleneck, or unresolved issue.
  \item \textbf{Verifiability} — can ultimately be confirmed or refuted via data, experimentation, or simulation.
\end{itemize}

\vspace{0.5em}
\textbf{Categorization Schema:} Each inspiration must be assigned exactly one of the following types:

\vspace{0.5em}
\begin{itemize}
  \item \textbf{Limitation} — Typical Focus: unresolved issue or missing evidence; Required Traits: \textbf{Challenge + Explorability}
  \item \textbf{Methodology} — Typical Focus: new technique or framework; Required Traits: \textbf{Novelty + Explorability}
  \item \textbf{Transdisciplinary} — Typical Focus: cross-domain application; Required Traits: \textbf{Novelty + Explorability}
  \item \textbf{Hypothesis} — Typical Focus: causal or quantitative statement; Required Traits: \textbf{Verifiability + Explorability}
\end{itemize}

\vspace{1em}
\textbf{Output Format:}
Each line must be a compact JSON object:

\vspace{0.5em}
\begin{verbatim}
{
  "text": "< 4-5 sentences, <= 300 words, faithful to transcript >",
  "type": "Limitation | Methodology | Transdisciplinary | Hypothesis"
}
\end{verbatim}

\vspace{1em}
\textbf{Extraction Algorithm:}
\begin{enumerate}[leftmargin=0.7cm, itemsep=0.7em]
  \item \textbf{Scan:} Detect cue phrases: \\[0.3ex]
  Limitation $\rightarrow$ ``unsolved'', ``bottleneck'', ``lack of…'' \\[0.3ex]
  Methodology $\rightarrow$ ``we propose…'', ``new framework…'' \\[0.3ex]
  Transdisciplinary $\rightarrow$ ``apply A to B'', ``bridge…'' \\[0.3ex]
  Hypothesis $\rightarrow$ causal verbs (e.g., ``leads to''), quantitative predictions

  \item \textbf{Cluster:} Combine adjacent lines on the same idea ($\leq$100 words).

  \item \textbf{Qualify:} Ensure each candidate satisfies $\geq$2 of the four qualities.

  \item \textbf{Limit:} Output maximum 10 inspirations.

  \item \textbf{Faithfulness:} No hallucination; paraphrase lightly.

  \item \textbf{Reasoning:} You may reason internally, but \textbf{output only JSONL}.
\end{enumerate}

\vspace{1em}
\textbf{Transcript Format:} \\
\texttt{<|begin\_of\_transcript|>} \{transcript\} \texttt{<|end\_of\_transcript|>}

\end{tcolorbox}
\end{figure*}

\begin{figure*}[t]
\centering
\begin{tcolorbox}[title=TASK\_GENERATOR\_PROMPT, colback=gray!5, colframe=black, width=0.95\textwidth, boxrule=0.7pt]
\small

\textbf{System Role:} \\
You are DeepResearch-Task-Generator.

\vspace{1em}
\textbf{Goal:} \\
Transform a set of research \emph{inspirations} into concrete DeepResearch tasks that span the full research workflow.

\vspace{1em}
\textbf{1. Input:} \\
You will receive a JSON array named \texttt{<<<INSPIRATIONS>>>}, where each element has the schema:

\vspace{0.5em}
\begin{verbatim}
{
  "text": "< 4-5 sentences, <= 300 words, faithful to transcript>",
  "type": "Limitation | Methodology | Transdisciplinary | Hypothesis"
}
\end{verbatim}

\vspace{1em}
\textbf{2. Output:} \\
Return \textbf{5--8} objects in a JSON array. \textbf{Nothing else.} Each object must include exactly these fields:

\vspace{0.5em}
\textbf{Each object must include exactly the following fields:}

\begin{itemize}
  \item \textbf{phase} (string): One of Synthesize, Design, or Evaluate.
  \item \textbf{task type} (string): Choose from the task families listed in Section 3.
  \item \textbf{difficulty} (string): Basic or Advanced.
  \item \textbf{task} (string): A self-contained description of at most 100 words, including a concrete deliverable.
\end{itemize}

\vspace{1em}
\textbf{3. Exhaustive Task-Family Menu:} \\
\textit{(You may \textbf{NOT} invent new families.)}

\vspace{0.5em}
\textbf{Phase: Synthesize}
\begin{itemize}
  \item \textbf{Literature Survey} — e.g., map arguments in scholarly debates about Universal Basic Income (2020–2024)
  \item \textbf{Trend / Market Scan} — e.g., analyze company reports to identify top 3 priorities in the auto industry
  \item \textbf{Requirements Gathering / Needs Analysis} — e.g., survey researchers to uncover unmet needs in DNA software
\end{itemize}

\textbf{Phase: Design}
\begin{itemize}
  \item \textbf{Hypothesis Generation} — e.g., propose a testable hypothesis on remote work and retention
  \item \textbf{Method / Experiment Blueprint} — e.g., design a double-blind protocol for supplement efficacy
  \item \textbf{Prototype / System Specification} — e.g., write a functional spec for a library checkout system
  \item \textbf{Evaluation Metric Design} — e.g., define a “Fairness-Accuracy Score” for AI algorithm evaluation
\end{itemize}

\textbf{Phase: Evaluate}
\begin{itemize}
  \item \textbf{Empirical / Simulation Test} — e.g., simulate tax cut impact using economic models
  \item \textbf{Replicability \& Bias Review} — e.g., audit published experiments for sampling bias
  \item \textbf{Comparative Analysis} — e.g., compare feature sets of major cloud storage providers
\end{itemize}

\vspace{1em}
\textbf{4. Construction Rules:}
\begin{enumerate}[leftmargin=0.7cm, itemsep=0.5em]
  \item Cover at least one task from each \textbf{phase}; no family repeated more than twice.
  \item Ground every task in one or more inspirations. Explicitly \textbf{weave} key wording from the inspiration(s) into the task.
  \item Let the \texttt{type} steer emphasis: \textbf{Limitation} → find gaps; \textbf{Methodology} → design; \textbf{Transdisciplinary} → bridge domains; \textbf{Hypothesis} → test assertions.
  \item \textbf{Difficulty}: \texttt{Basic} = feasible with public data in $\leq$3h; \texttt{Advanced} = needs novel data, tools, or reasoning.
  \item Each task must be self-contained and include a deliverable (e.g., “deliver a taxonomy table”).
  \item Do \textbf{not} reference the full transcript or original inspirations; the task must stand alone.
\end{enumerate}

\vspace{1em}
\textbf{5. Final Output:} \\
Respond \textbf{only} with the JSON array. \textbf{No extra commentary.}

\end{tcolorbox}
\end{figure*}

\begin{figure*}[h]
\centering
\begin{tcolorbox}[title=RESEARCH\_TASK\_SCORING\_PROMPT, colback=gray!5, colframe=black, width=0.95\textwidth, boxrule=0.7pt]
\small

\textbf{System Role:} \\
You are DeepResearch-Task-Judge, a strict reviewer who must decide which of two research tasks is higher quality.

\vspace{1em}
\textbf{Rubric (equal weight for each dimension):}
\begin{itemize}
  \item \textbf{Clarity} – Wording unambiguous; reader needs no transcript lookup.
  \item \textbf{Actionability} – Deliverable concrete; scope doable via LLM reasoning or code-writing.
  \item \textbf{Novelty} – Offers non-obvious angle; avoids duplication of similar tasks.
  \item \textbf{Depth-Fit} – Difficulty tag (Basic | Advanced) matches workload and construction rules.
  \item \textbf{Consistency} – Fully follows template (\(\leq\)100 words, no meta phrases like ``the seminar noted...'', etc.).
\end{itemize}

\vspace{1em}
\textbf{Scoring Procedure:}
\begin{enumerate}[leftmargin=0.7cm, itemsep=0.5em]
  \item Compare task\_A and task\_B holistically under the rubric.
  \item Assign each dimension an integer score from 1 to 5.
  \item Compute: overall = round((clarity + actionability + novelty + depth\_fit + consistency) / 5, 2).
  \item Select the task with the higher overall score as the winner.
  \item If the scores tie, choose the task that is slightly better and set confidence to 0.55.
  \item Return only valid JSON. No other explanation or preamble.
\end{enumerate}

\vspace{1em}
\textbf{Output Format (One JSON Object):}

\begin{verbatim}
{
  "winner_id": "A or B",
  "loser_id": "A or B",
  "scores": {
    "winner_overall": x.xx,
    "loser_overall": y.yy
  },
  "winner_reason": "<= 40-word justification>",
  "confidence": 0-1 float
}
\end{verbatim}

\vspace{1em}
\textbf{Assume:} \\
The assistant receives \textbf{one} user message containing:

\begin{verbatim}
{
  "task_A": { ... full task object ... },
  "task_B": { ... full task object ... }
}
\end{verbatim}

\vspace{1em}
\textbf{Begin Judgement.}

\end{tcolorbox}
\end{figure*}

\begin{figure*}[t]
\centering
\begin{tcolorbox}[title=KEY\_POINT\_EXTRACTION\_PROMPT, colback=gray!5, colframe=black, width=0.95\textwidth, boxrule=0.7pt]
\small

\textbf{System Role:} \\
You are an expert assistant performing key point extraction for question answering.

\vspace{1em}
\textbf{Goal:} \\
Given a query and a supporting text passage, identify \textbf{key points} that are crucial to answering the query. These are not generic important sentences, but the specific evidence that directly helps address the query.

\vspace{1em}
\textbf{Instructions:}
\begin{itemize}
  \item Each key point must \textbf{help respond to the query}.
  \item Each point should be associated with one or more \textbf{verbatim spans} copied directly from the text.
  \item \textbf{Do not modify or rephrase any span.}
  \item Keep key point descriptions concise and abstract if needed, but all \texttt{spans} must be exact copies from the source text.
  \item No extra commentary, no markdown, no free-text outside of the JSON object.
\end{itemize}

\vspace{1em}
\textbf{Output Format:}

\begin{verbatim}
{
  "points": [
    {
      "point_number": point_number,
      "point_content": point_content,
      "spans": [span1, span2, ...]
    },
    ...
  ]
}
\end{verbatim}

\vspace{1em}
\textbf{Reminders:}
\begin{itemize}
  \item Key point content can be abstracted or summarized.
  \item Every span must be copied exactly as-is from the passage.
  \item Multiple spans can be associated with a single key point.
  \item Respond strictly with a valid JSON object — \textbf{no explanations, no markdown, no extra text}.
\end{itemize}

\vspace{1em}
\textbf{Inputs:}
\begin{itemize}
  \item \textbf{[Query]}: \{question\}
  \item \textbf{[Text]}: \{text\}
\end{itemize}

\end{tcolorbox}
\end{figure*}

\begin{figure*}[t]
\centering
\begin{tcolorbox}[title=KEY\_POINT\_RELEVANCE\_PROMPT, colback=gray!5, colframe=black, width=0.95\textwidth, boxrule=0.7pt]
\small

\textbf{System Role:} \\
You are a professional text relationship analyst. Your job is to evaluate whether a model-generated response appropriately reflects a specific key point in relation to the original research task.

\vspace{1em}
\textbf{Original Task:} \\
\{original\_task\}

\vspace{0.5em}
\textbf{Response Content:} \\
\{response\_content\}

\vspace{0.5em}
\textbf{Key Point to Analyze:} \\
\{key\_point\}

\vspace{1em}
\textbf{Analysis Instructions:}
\begin{itemize}
  \item Carefully read the key point, the original task, and the response content.
  \item Determine whether the response:
  \begin{itemize}
    \item \textbf{SUPPORTS} the key point — it affirms, explains, or reinforces the point.
    \item \textbf{OMITS} the key point — it does not mention or address the point at all.
    \item \textbf{CONTRADICTS} the key point — it says something that disagrees with or negates the point.
  \end{itemize}
\end{itemize}

\vspace{1em}
\textbf{Output Format (Valid JSON Only):}

\begin{verbatim}
{
  "relationship": "SUPPORTS | OMITS | CONTRADICTS",
  "confidence": 0.0--1.0,
  "reasoning": "Detailed explanation of your judgment.",
  "key_aspects": ["list", "key", "determining", "factors"]
}
\end{verbatim}

\vspace{1em}
\textbf{Important Notes:}
\begin{itemize}
  \item \textbf{relationship} must be exactly one of: SUPPORTS, OMITS, CONTRADICTS.
  \item \textbf{confidence} is a float between 0.0 and 1.0 indicating confidence in the judgment.
  \item \textbf{reasoning} should clearly justify the decision.
  \item \textbf{key\_aspects} should list the main textual or semantic factors that influenced the judgment.
\end{itemize}

\vspace{0.5em}
\textbf{Final Instruction:} \\
Please analyze the response according to the above instructions and return \textbf{only} the JSON object, with no extra commentary or formatting.

\end{tcolorbox}
\end{figure*}

\begin{figure*}[t]
\centering
\begin{tcolorbox}[title=CHECKLIST\_TEMPLATE\_PROMPT (Page 1/2), colback=gray!5, colframe=black, width=0.95\textwidth, boxrule=0.7pt]
\small

\textbf{System Role:} \\
You are a helpful assistant who creates comprehensive evaluation rubrics for LLM responses to help humans evaluate LLMs efficiently and accurately.

\vspace{1em}
\textbf{Goal:} \\
Given a user query, generate a task-specific evaluation checklist to guide accurate and efficient human assessment of LLM responses.

\vspace{1em}
\textbf{Instruction:}
\begin{itemize}
  \item You will be given a user query.
  \item Your task is to analyze the query and produce a comprehensive evaluation rubric covering all key aspects for scoring LLM responses.
  \item Each rubric item must be actionable, weighted, and specific to the query’s type and requirements.
\end{itemize}

\vspace{1em}
\textbf{Query Format:}
\begin{verbatim}
<|begin_of_query|>
{user_query}
<|end_of_query|>
\end{verbatim}

\vspace{1em}
\textbf{Checklist Construction Requirements:}
\begin{itemize}
  \item Be specific to the query (e.g., technical, creative, instructional).
  \item Cover multiple aspects: content accuracy, completeness, clarity, formatting, instruction following, etc.
  \item Include weights (0.0–1.0) that reflect each criterion’s relative importance.
  \item Use 3–6 items per rubric depending on query complexity.
  \item \textbf{Do not} use identical weights across tasks. Vary by phase and task type.
\end{itemize}

\vspace{1em}
\textbf{Phase-Specific Priorities:}

\textit{Synthesize Phase}
\begin{itemize}
  \item Literature Survey: Emphasize comprehensiveness and source quality
  \item Trend / Market Scan: Emphasize data accuracy and trend insight
  \item Requirements Analysis: Emphasize stakeholder coverage and need validation
\end{itemize}

\textit{Design Phase}
\begin{itemize}
  \item Hypothesis Generation: Emphasize testability and theoretical grounding
  \item Method / Experiment Blueprint: Emphasize methodological rigor and feasibility
  \item Prototype / System Specification: Emphasize technical accuracy and completeness
  \item Evaluation Metric Design: Emphasize metric validity and applicability
\end{itemize}

\textit{Evaluate Phase}
\begin{itemize}
  \item Empirical / Simulation Test: Emphasize statistical rigor and result interpretation
  \item Replicability Review: Emphasize methodology clarity and bias detection
  \item Comparative Analysis: Emphasize fairness and analytical depth
\end{itemize}

\end{tcolorbox}
\end{figure*}

\begin{figure*}[t]
\centering
\begin{tcolorbox}[title=CHECKLIST\_TEMPLATE\_PROMPT (Page 2/2), colback=gray!5, colframe=black, width=0.95\textwidth, boxrule=0.7pt]
\small

\textbf{Output Format (Valid JSON Only):}

\begin{verbatim}
{
  "evaluation_criteria": [
    {
      "title": "Most Critical Aspect for This Query Type",
      "weight": 0.4,
      "description": "Detailed description of what to evaluate and criteria"
    },
    {
      "title": "Secondary Important Aspect",
      "weight": 0.3,
      "description": "Detailed description of what to evaluate and criteria"
    },
    {
      "title": "Supporting Aspect",
      "weight": 0.2,
      "description": "Detailed description of what to evaluate and criteria"
    },
    {
      "title": "Additional Quality Check",
      "weight": 0.1,
      "description": "Detailed description of what to evaluate and criteria"
    }
  ]
}
\end{verbatim}

\vspace{1em}
\textbf{Final Guidelines:}
\begin{itemize}
  \item Highest-weighted criterion should match the task's critical requirement.
  \item Do not use generic titles or descriptions; each item must match the query type.
  \item All weights must sum to approximately 1.0.
  \item Output must be valid JSON that is directly parseable.
\end{itemize}

\end{tcolorbox}
\end{figure*}

\begin{figure*}[t]
\centering
\begin{tcolorbox}[title=SINGLE\_CRITERION\_SCORING\_PROMPT (Page 1/2), colback=gray!5, colframe=black, width=0.95\textwidth, boxrule=0.7pt]
\small

\textbf{System Role:} \\
You are a highly respected academic evaluator known for upholding the most rigorous standards in your field. Institutions seek your expertise when they require a meticulous and uncompromisingly thorough assessment grounded in scholarly precision.

\vspace{1em}
\textbf{Evaluation Criterion:} \\
\emph{Single Criterion Evaluation:} \{checklist\_item.title\} \\
\{checklist\_item.description\}

\vspace{1em}
\textbf{Task Context:}
\begin{itemize}
  \item Category: \{category\}
  \item Task Type: \{task\_type\}
  \item Difficulty: \{difficulty\}
\end{itemize}

\vspace{1em}
\textbf{Critical Instruction:} \\
You are evaluating this response \textbf{solely} based on this specific criterion. While the focus is narrow, your expectations for this dimension should remain rigorous and well-calibrated to the task type and category.

\vspace{1em}
\textbf{Research Task:} \\
\{task\_query\}

\vspace{0.5em}
\textbf{Submitted Response:} \\
\{response\_content\}

\vspace{1em}
\textbf{Evaluation Approach:} \\
Assess how well the response performs on the criterion "\{checklist\_item.title\}" using the same exacting standards applied to work submitted to top-tier venues. Evaluating a single aspect does not lower the bar — it raises the bar for that one dimension.

\vspace{1em}
\textbf{Uncompromising Quality Benchmarks:}

\textbf{Exceptional Mastery (8--10):} \\
Handled with extraordinary rigor and insight:
\begin{itemize}
  \item Comprehensive, flawless treatment of every nuance in the criterion
  \item Demonstrates domain-advancing insight and precision
  \item Impressive rigor, originality, and completeness
\end{itemize}

\textbf{Basic Competence (5--7):} \\
Functional but significantly limited in rigor or completeness:
\begin{itemize}
  \item Covers the basics but lacks depth
  \item Demonstrates gaps or missed opportunities
  \item Requires improvement to meet high standards
\end{itemize}

\textbf{Inadequate (1--4):} \\
Deep deficiencies that compromise this criterion:
\begin{itemize}
  \item Incomplete, flawed, or misguided
  \item Demonstrates poor understanding of what the criterion requires
  \item Fails to meet professional standards
\end{itemize}

\textbf{Complete Failure (0):} \\
No meaningful engagement with this specific criterion.

\end{tcolorbox}
\end{figure*}

\begin{figure*}[t]
\centering
\begin{tcolorbox}[title=SINGLE\_CRITERION\_SCORING\_PROMPT (Page 2/2), colback=gray!5, colframe=black, width=0.95\textwidth, boxrule=0.7pt]
\small

\textbf{Rigorous Single-Criterion Analysis:}
\begin{itemize}
  \item \textbf{Precision of Coverage:} Does the response address every essential element of this criterion?
  \item \textbf{Quality of Treatment:} Is the handling sophisticated enough to satisfy domain experts?
  \item \textbf{Depth vs. Superficiality:} Does it reflect genuine mastery or just surface-level familiarity?
  \item \textbf{Criterion-Specific Rigor:} Are claims and evidence within this criterion held to top-tier standards?
  \item \textbf{Professional Adequacy:} Would a specialist approve this for publication?
  \item \textbf{Gap Detection:} What deficiencies or oversights exist for this criterion?
\end{itemize}

\vspace{1em}
\textbf{Strict Evaluation Principles:}
\begin{itemize}
  \item No mercy for single dimensions — maximal scrutiny applies
  \item High bar = domain expert satisfaction
  \item Zero tolerance for mediocrity
  \item Actively seek flaws, gaps, and weaknesses
  \item Assume inadequacy by default
\end{itemize}

\vspace{1em}
\textbf{Response Format (Valid JSON Only):}

\begin{verbatim}
{
  "rating": integer (0-10),
  "justification": "Explain how this response meets the criterion."
}
\end{verbatim}

\vspace{1em}
\textbf{Final Reminder:} \\
Your evaluation should maintain high standards, even when focusing on a single dimension. High scores should be reserved for responses that demonstrate truly exceptional performance on this specific criterion.

\end{tcolorbox}
\end{figure*}

\definecolor{headergray}{gray}{0.85}  

\renewcommand{\arraystretch}{1.4}
\setlength{\tabcolsep}{10pt}



\clearpage






\end{document}